\documentclass[10pt, twoside, twocolumn]{IEEEtran}
\usepackage{cite}      
\usepackage{bm}
\usepackage{makeidx}  
\usepackage{times}
\usepackage{graphicx}
\usepackage{subfigure}
\usepackage{threeparttable}
\usepackage{amsmath}
\usepackage{amssymb}
\usepackage{color}
\begin{document}

\title{Unsupervised Feature Learning with C-SVDDNet}

\author{Dong~Wang and Xiaoyang~Tan \thanks{Dong Wang and Xiaoyang Tan are with the
    Department of Computer Science and Technology, Nanjing University
    of Aeronautics and Astronautics, P.R.~China. Corresponding author: Xiaoyang Tan
    (x.tan@nuaa.edu.cn).}}

\maketitle

\begin{abstract}
In this paper, we investigate the problem of learning feature representation from unlabeled data using a single-layer K-means network. A K-means network maps the input data into a feature representation by finding the nearest centroid for each input point, which has attracted researchers' great attention recently due to its simplicity, effectiveness, and scalability. However, one drawback of this feature mapping is that it tends to be unreliable when the training data contains noise. To address this issue, we propose a SVDD based feature learning algorithm that describes the density and distribution of each cluster from K-means with an SVDD ball for more robust feature representation. For this purpose, we present a new SVDD algorithm called C-SVDD that centers the SVDD ball towards the mode of local density of each cluster, and we show that the objective of C-SVDD can be solved very efficiently as a linear programming problem. Additionally, traditional unsupervised feature learning methods usually take an average or sum of local representations to obtain global representation which ignore spatial relationship among them. To use spatial information we propose a global representation with a variant of SIFT descriptor. The architecture is also extended with multiple receptive field scales and multiple pooling sizes. Extensive experiments on several popular object recognition benchmarks, such as STL-10, MINST, Holiday and Copydays shows that the proposed C-SVDDNet method yields comparable or better performance than that of the previous state of the art methods.

\end{abstract}





%
\section{Introduction}\label{introduction}
Learning good feature representation from unlabeled data is the key to make progress in recognition and classification tasks, and has attracted great attention and interest from both academia and industry recently. A representative method for this is the deep learning (DL) approach \cite{bengio2012representation} with its goal to learn multiple layers of abstract representations from data. Among others, one typical DL method is the so called convolutional neural network (ConvNet), which consists of multiple trainable stages stacked on top of each other, followed by a supervised classifier \cite{lecun1998gradient} \cite{jarrett2009best}. Many variations of ConvNet network have been proposed as well for different vision tasks \cite{bruna2013invariant}\cite{le2011building}\cite{carneiro2012segmentation}\cite{San2014Evolvable} \cite{Shuai2014Nonlinearly} with great success.

In these methods layers of representation are usually obtained by greedily training one layer at a time on the lower level \cite{le2011building} \cite{agarwal2006hyperfeatures} \cite{jarrett2009best}, using an unsupervised learning algorithm. Hence the performance of single-layer learning has a big effect on the final representation. Neural network based single-layer methods, such as autoencoder \cite{hinton2006reducing} and  RBM (Restricted Boltzmann Machine,\cite{cueto2010geometry}), are widely used for this but they usually have many parameters to adjust, which is very time-consuming in practice.

That motivates more simple and more efficient methods for single-layer feature learning. Among others K-means clustering algorithm is a commonly used unsupervised learning method, which maps the input data into a feature representation simply by associating each data point to its nearest cluster center. There is only one parameter involved in the K-means based method, i.e., the number of clusters, hence the model is very easy to use in practice. Coates et al. \cite{coates2010analysis} shows that the K-means based feature learning network is capable to achieve superior performance compared to sparse autoencoder, sparse RBM and GMM (Guassian Mixture Model). However, the K-means based feature representation may be too terse, and does not take the non-uniform distribution of cluster size into account - Intuitively, clusters containing more data are likely to be part of the features with higher influential power, compared to the smaller ones.

In this paper, we proposed a SVDD (Support Vector Data Description,  \cite{tax2004support}, \cite{xu2011fault}, \cite{banerjee2010efficient}) based method to address these issues. The key idea of our method is to use SVDD to measure the density of each cluster resulted from K-means clustering, based on which more robust feature representation is built. Actually the K-means algorithm lacks a robust definition of the size of its clusters, since the nearest center principle is not robust against the noise or outliers commonly encountered in real world applications. We advocate that the SVDD could be a good way to address this issue. Actually SVDD is a widely used tool to find a minimal closed spherical boundary to describe the data belonging to the target class and therefore, given a cluster of data, we are expecting SVDD to generate a ball containing the normal data except outliers. Performing this procedure on all the clusters of K-means, we will finally obtain $K$ SVDD balls on which our representation can be built. In addition, to take the cluster size into account, we use the distance from the data to each ball's surface instead of the center as the feature.

One possible problem of this method, however, may come from the instability of SVDD's center, due to the fact that its position is mainly determined by the support vectors on the boundary and the noise in the data may deviate the center far from the mode (c.f., Fig.~\ref{fig:LPSVDD} (left)). Hence the resulting SVDD ball may not be consistent with the data's distribution when used for feature representation. To address this issue, we add a new constraint to the original SVDD objective function to make the model align better with the data. In addition, we show that our modified SVDD can be solved very efficiently as a linear programming problem, instead of as a quadratic one. Usually we need to compute hundreds of clusters, and a linear programming solution can thus save us large amounts of time. The proposed method is further extended by adopting a set of receptive fields with different sizes to capture multi-scale information ranging from detailed edge-like features to part-level features. A preliminary version of this work appeared in \cite{wang2013centering}, and the feasibility and effectiveness of the proposed C-SVDD-based method (called C-SVDDNet) is verified extensively on several object recognition and image retrieval benchmarks with competitive performance.

The remaining parts of this paper are organized as follows: In Section~\ref{sec_prelimiaries}, preliminaries are provided regarding unsupervised feature learning representation, then we detail our improved feature learning method in Section~\ref{sec_proposedmethod}. In Section~\ref{sec_exps}, we investigate the performance of our method empirically over several popular datasets. We conclude this paper in Section~\ref{sec_conclude}.

\section{Unsupervised Feature Learning }\label{sec_prelimiaries}
The goal of unsupervised feature learning is to automatically discover useful hidden patterns/features in large datasets without relying on a supervisory signal, and those learnt patterns can be utilized to create representations that facilitate subsequent supervised learning (e.g., object classification). Compared to supervised learning, unsupervised learning has its unique characteristics and advantages. Among others, one of the most important one is that it can be used to learn consistent patterns from unlabelled data, which are often free and easy to obtain. Such patterns distinguish from noise since by definition noise can be thought of as random variations presented in the data. This implies many potential applications of unsupervised learning, e.g., to transfer knowledge from one domain to another related domain, to regularize the behavior of a supervised algorithm, and to represent the data in a compact but effective manner. Due to these reasons, unsupervised learning are regarded as the future of deep learning \cite{Lecun2015nature}.
%

There are many kinds of unsupervised learning methods in computer vision, such as Bag of Words (BoW) \cite{csurka2004visual}, Vector of Linearly Aggregated Descriptors (VLAD) \cite{jegou2012aggregating}, Fisher vector (FV) \cite{sanchez2013image}, and so on. A typical pipeline for unsupervised feature learning includes three steps. The first step is to train a set of local filters from the unlabeled training data. This is usually done by running K-means (for BoW, VLAD) or GMM (for FV) on lots of local patches sampled from the dataset and then using the centers of clusters as filter bank. The second step is to partition a given image into patches and encode them into a set of feature vectors using the learnt filter bank. These feature vectors are finally combined and normalized as the feature representation for the input image. In what follows we give a brief review on these methods.

\def\para#1{\medskip\noindent{\bf #1}}
\para{Bag of Words and Its Variants} The simple and basic unsupervised feature learning method is the BoW model. In this model local filters are usually the centers of clusters from K-means. These filters are looked as bins, which serves to pool the local patches nearest to them. This can be regarded as a "hard voting" method:
\begin{equation}\label{eq_kmeanscoding}
f_{k}(x) = \left\{ \begin{array}{ll}
1 \ & \textrm{if $k=argmin_{j}\Arrowvert c_{j}-x \Arrowvert_{2}^2 $}\\
0 \ & \textrm{otherwise}
\end{array} \right.
\end{equation}
where $f_{k}(x)$ is the value that a patch $x$ was encoded as with the k-th filter $c_k$. In BoW, we simply count the number of patches in each bin to get a histogram representation. Thus it is a very coarse way to encode the information of an input image.

Alternatively, VLAD \cite{jegou2012aggregating} and FV \cite{sanchez2013image} encode each data point $x$ with a vector instead of a simple count number as in BoW, which effectively improve the richness and robustness of the feature representation. Particularly, FV captures the first and second order difference between an input $x$ and the centres of a GMM, denoted as $c_k$,
\begin{equation}\label{eq_fv}
\begin{split}
 &f_{uk}(x) = \frac{1}{N\sqrt{\pi_k}}q_{ik}\Sigma_k^{-\frac{1}{2}}(x-c_k) \\
 &f_{vk}(x) = \frac{1}{N\sqrt{2\pi_k}}q_{ik}[(x-c_k)\Sigma_k^{-\frac{1}{2}}(x-c_k)-1] \\
\end{split}
\end{equation}
Then the FV coding for an local patch $x$ is a vector of $[f_{u1}^T(x),f_{v1}^T(x),f_{u2}^T(x),f_{v2}^T(x),...,f_{uK}^T(x),f_{vK}^T(x)]^T$. VLAD \cite{jegou2012aggregating} is a simplified version of FV, with the difference signal between a patch $x$ and a filter $c_k$ defined as $f_{k}(x) = x - c_k$. As in FV, these difference signals are concatenated into a $K$-dim vector for feature representation. Obviously, both FV and VLAD encode much richer information than that of BoW, hence being more discriminative in subsequent tasks such as object classification.

\para{Coates et al.'s Method} To the best of our knowledge, the work of \cite{coates2010analysis} is the first ``deep" unsupervised learning method that are based on the K-means method, hence having close connection with the aforementioned BoW, VLAD and FV methods. Particularly, after learning a filter bank, instead of using it as basin of attraction like in BoW or as references for calculating difference vectors, it is utilized to generate a series of feature maps, one for each filter. This has at least two potential advantages: 1) compared to VLAD and FV, the encoded information is even more rich; 2) the feature maps preserve the spatial information well and hence the whole procedure could be repeated, leading to a deep unsupervised learning architecture.

Furthermore, to deal with the problem of ``hard coding" in K-means, the following ``triangle" encoding is proposed \cite{coates2010analysis}:
\begin{equation}\label{eq_soft}
f_{k}(x) = max\{0,\mu(z)-z_{k}(x)\}
\end{equation}
where $z_{k}(x)=\Arrowvert x-c_{k} \Arrowvert_{2}$, and $\mu(z)$ is the mean of the elements of $z$. This activation function outputs $0$ for the feature $f_{k}$ that has an ¡°above average¡± distance to the centroid $c_{k}$. This model leads to a less sparse representation (roughly half of the features could be set to be $0$). Note that this ``triangle" encoding strategy essentially allows us to learn a distributed representation using the simple K-means method instead of more complicated network-based methods (e.g., autoencoder and RBM), hence saving much time in training. Coastes et al. \cite{coates2010analysis} shows that this strategy actually leads to comparable performance to, if not better than, those based on network methods.

\begin{figure}[t]
\centering
\includegraphics[width=0.8\linewidth]{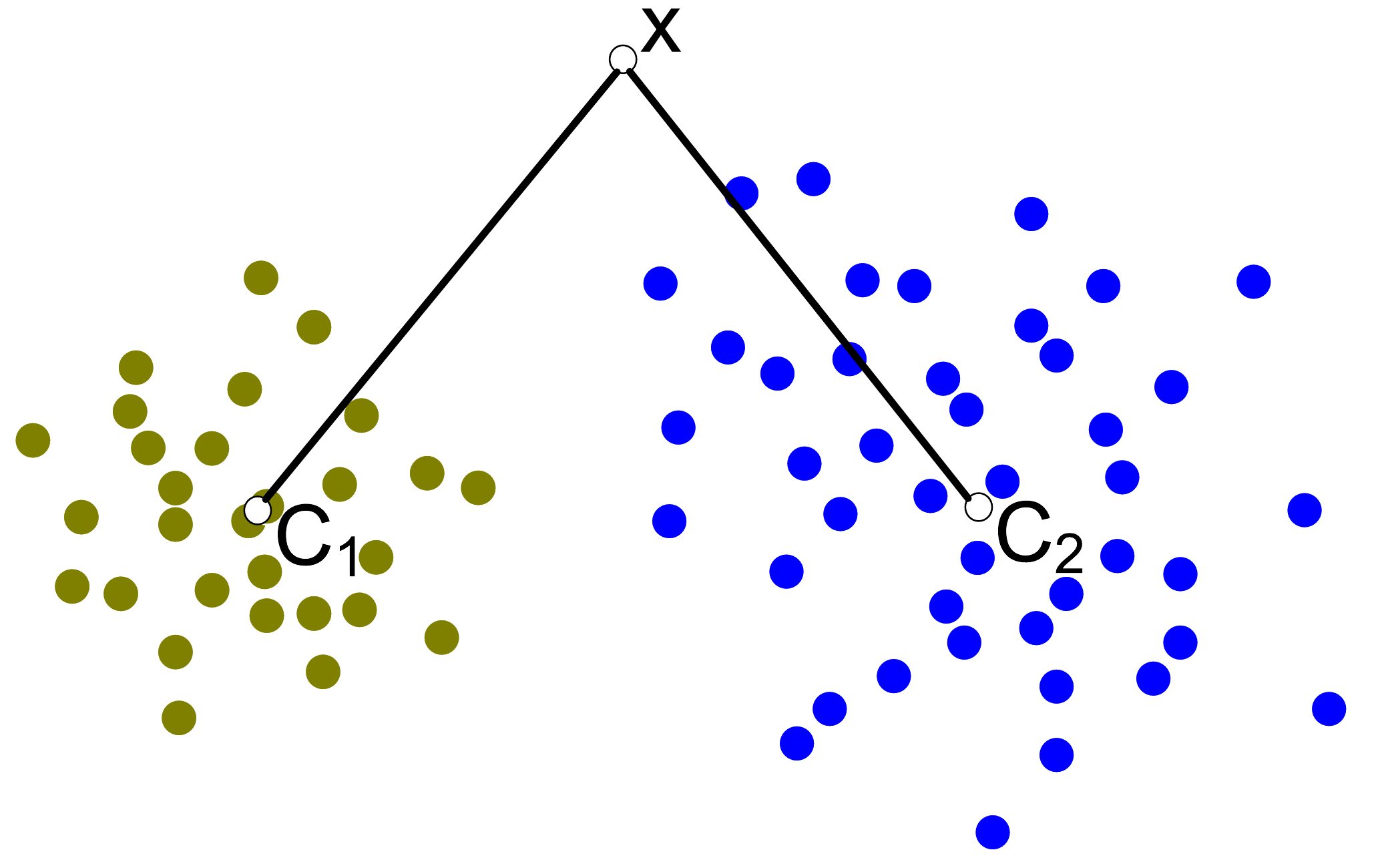}
\caption{Illustration of the unequal cluster effect, where the distances from a test point $x$ to two cluster centers $C_1$ and $C_2$ are equal but the size of two clusters are different.}
\label{fig:unequal}
\end{figure}

However, this method does not take the characteristics of each cluster into consideration. Actually, the number of data point in each cluster is usually different, so is the distribution of data points in each cluster. We believe that these differences would make a difference in feature representation as well. Unfortunately the aforementioned K-means feature mapping scheme completely ignores these and only uses the position of center for feature encoding. As shown in Fig.~\ref{fig:unequal}, although the data point $x$ has the same distance to the centers $C_1$ and $C_2$ of two clusters, it should be assigned a different score on $C_1$ than on $C_2$ since the former cluster $C_1$ is much bigger than the latter. In practice such unequal clusters are not uncommon and the K-means method by itself can not reliably grasp the size of its clusters due to the existence of outliers. To this end, we propose a SVDD based method to describe the density and distribution of each cluster and use this for more robust feature representation.

\section{The Proposed Method}\label{sec_proposedmethod}

In this section, after presenting an overview of the proposed method, we give the details of our Centered-SVDD method for feature encoding, and compare it with the K-means ``triangle" encoding method. Then we describe our SIFT-based post-pooling layer and discuss how to extend the method to extract multi-scale information.

\subsection{Overview of the Proposed Method}
A typical single-layer network contains several components: an input image is first mapped into a set of feature maps using filter banks (or dictionary), which are then subjected to a pooling/subsampling operation to condense the information contained in the feature maps. Finally, the pooled feature maps are concatenated to a feature vector, which serves as the representation for the subsequent classification/cluster tasks. There are several design options in this procedure, where the size of filter bank and that of the pooling grids are the major tradeoff one has to make.

Generally speaking, bigger filter banks help each sample find its nearby representative points more accurately but at the cost of yielding a high-dimensional representation, hence a crude pooling/subsampling is needed to reduce the dimensionality. Overall this type of architecture emphasizes more on the global aspects of the samples than on the local ones (e.g., local texture, local shape, etc.). Actually, Coates et al. show that this kind of network is able to yield state of the art results on several challenging datasets \cite{coates2010analysis}. On the other hand, other works use smaller filter banks but highlight the importance of detailed local information in constructing the representation, usually based on some complicated feature encoding strategy, as done in PCANet \cite{chan2014pcanet} or Fisher Vector \cite{chatfield2011devil}.

In this work, we follow the second design choice, based on the consideration that the learned representation should preserve enough local spatial information for the subsequent processing. Compared to \cite{coates2010analysis}, we use an improved feature encoding method named C-SVDD (detailed in the next section) and adopt the architecture of relatively small dictionary. Different to  \cite{chan2014pcanet} or \cite{chatfield2011devil}, we learn filter banks for feature encoding but add a SIFT-based post-pooling processing procedure onto the network, which essentially projects the responses of a pooling operation into a more compact and robust representation space. 



\subsection{Using SVDD Ball to Cover Unequal Clusters}\label{sec_svddcoding}
Assume that a dataset contains $N$ data objects, \{$x_{i}\}$, $i=1,...,n$ and a ball is described by its center $a$ and the radius $R$. The goal of SVDD (Support Vector Data Description, \cite{tax2004support}) is to find a closed spherical boundary around the given data points. In order to avoid the influence of outliers, SVDD actually faces the tradeoff between two conflicting goals, i.e., minimizing the radius while covering as many data points as possible.
This can be formulated as the following objective,
\begin{equation}
\label{eq_svdd}
\begin{split}
 &min_{a, R,\xi_{i}}  \ R^2+\lambda\sum_{i=1}^{N}\xi_{i} \\
 &s.t. \ \Arrowvert x_{i}-a \Arrowvert^2 \le R^2+\xi_{i} \\
 & ~~~~~~ \xi_{i} \ge 0,
\end{split}
\end{equation}
where the slack variable $\xi$ represents the penalty related with the deviation of the $i$-th training data point outside the ball, and $\lambda$ is a user defined parameter controlling the degree of regularization imposed on the objective.
With the KKT conditions, we have $a=\sum_{i=1}^N x_i$, i.e., the center $a$ of the ball is a linear combination of the data $x_{i}$. The dual function of Eq.(~\ref{eq_svdd}) is
\begin{equation}
\begin{split}
 &max \ \sum_{i}\alpha_{i} \langle x_{i},x_{i} \rangle-\sum_{i}\sum_{j}\alpha_{i}\alpha_{j} \langle x_{i},x_{j}\rangle  \\
 &s.t. \ \sum_{i}\alpha_{i}=1 \ , \ \alpha_{i}\in[0,\lambda]\ ,\ i=1,...,N,
\end{split}
\end{equation}
where $\alpha_{i}$ and $\alpha_{j}$ are Lagrangian multipliers. By solving the quadratic programming problem we can get the center $a$ and the radius $R$.

The SVDD method can be understood as a type of one-class SVM and its boundary is solely determined by support vectors points. SVDD allows us to summarize a group of data points in a nice and robust way. Hence it is natural to use SVDD ball to model each cluster from K-means, thereby combining the strength of both models. In particular,  for a given data point we first compute its distance $h_k$ to the surface of each SVDD ball $C_k$, and then use the following modified ``triangle" encoding method for feature representation (c.f., E.q.(~\ref{eq_soft})),
\begin{equation}\label{eq_softsvdd}
f_{k}(x) = max\{0,g(h)-h_{k}(x)\},
\end{equation}
where $h_{k}(x)=||x-R_{k}||_2$ is the distance from the point $x$ to the surface of the $k$-th SVDD ball, while $g(h)$ is the average of the values $h_k$.

\begin{figure}[t]
\centering
\includegraphics[width=0.8\linewidth]{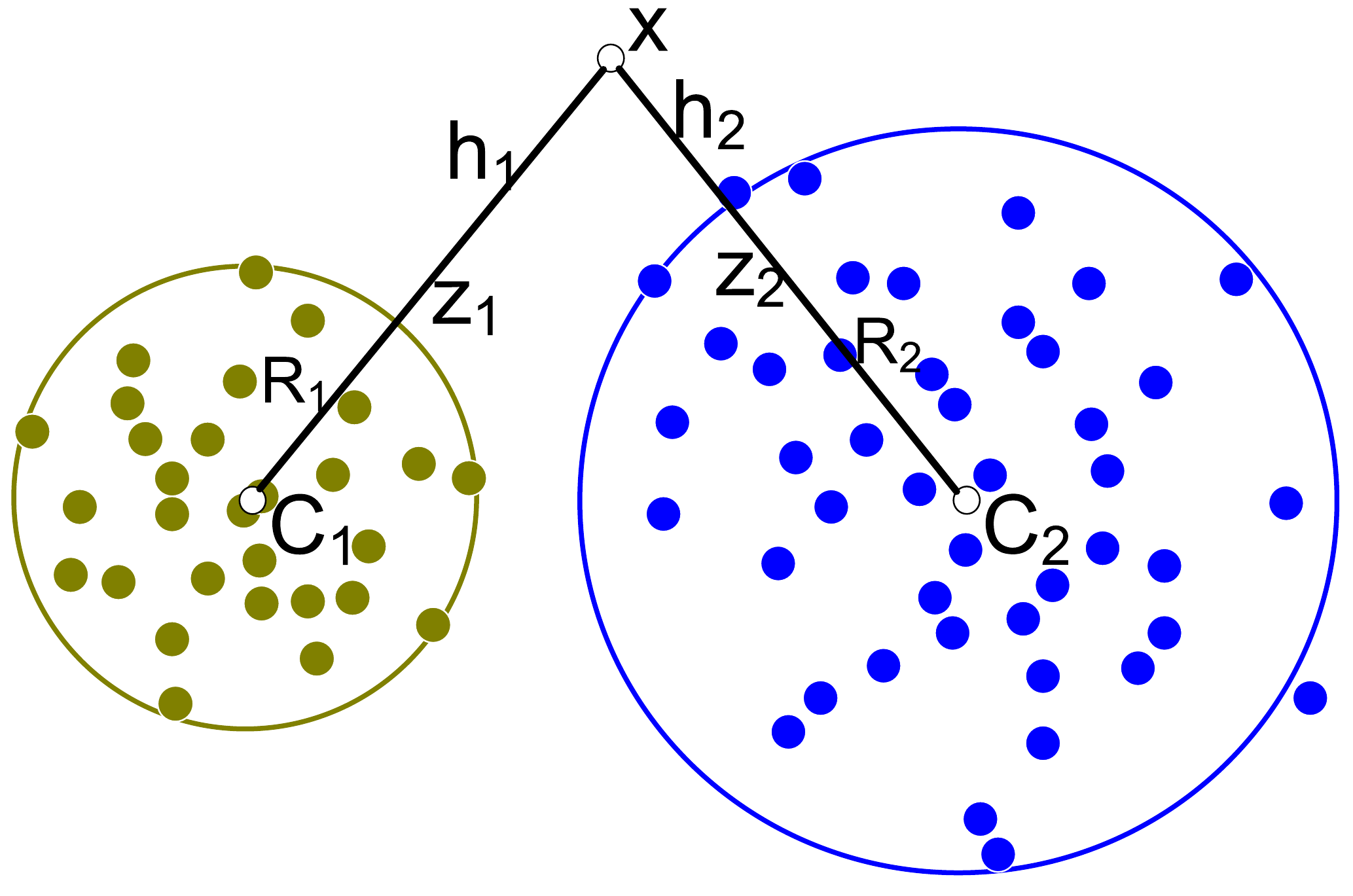}
\caption{Using the SVDD ball to cover the clusters of K-means, where two SVDD balls cover two clusters with different sizes, respectively. For a test point $x$, we encode its feature using its distance $h$ to the surface of an SVDD ball. This can be calculated by subtracting the length $R$ of the radius of the ball from the distance $z$ between $x$ to the ball center $C$. Hence for two SVDD balls with different size, the encoded features for the same point $x$ would be different.}
\label{fig:SVDD}
\end{figure}

Shown in Fig.~\ref{fig:SVDD} for a data point $x$, $C_{i}, {i=1,2}$ respectively are the centroids of two SVDD balls with $R_{i}, {i=1,2}$ being the radius. Since the distances from $x$ to $C_{1}$ and $C_{2}$ are equal, $x$ will be assigned the same scores on the two ball with the K-means scheme (c.f., E.q.(~\ref{eq_soft})). However, if we take the density and size of the clusters into accounts, the score from $C_2$ should be higher in our method.

\subsection{The C-SVDD Model}

\begin{figure}[t]
\centering
\includegraphics[width=1\linewidth]{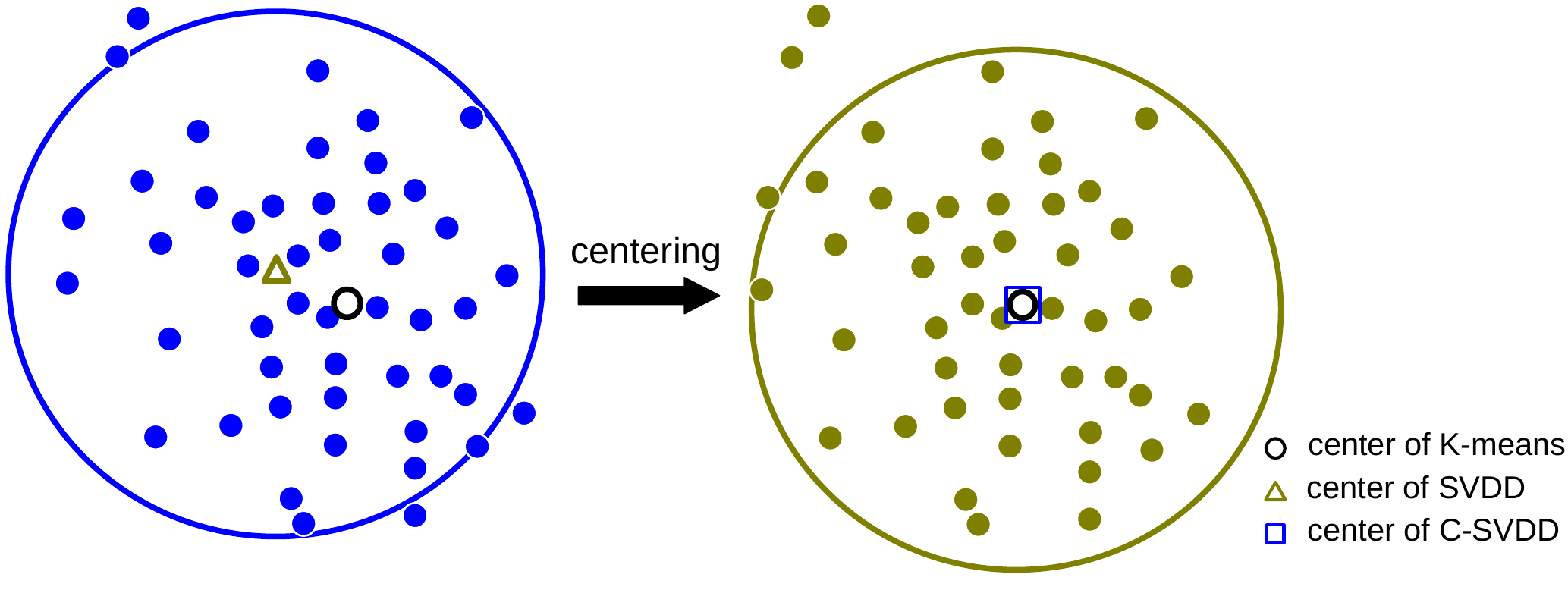}
\caption{Illustration of the difference between SVDD and C-SVDD. Note that after centering the SVDD ball (left), the center of C-SVDD ball (right) aligns better with the high density region of the data points.}
\label{fig:LPSVDD}
\end{figure}

Although SVDD ball provides a robust way to describe the cluster of data, one unwelcome property of the ball is that it may not align well with the distribution of data points in that cluster. As illustrated in Fig.~\ref{fig:LPSVDD} (left), although the SVDD ball covers the cluster $C_1$ well, its center is biased to the region with low density. This should be avoided since it actually gives suboptimal estimates on the distribution of the cluster of data.

To address this issue, inspired by the observation that the centers of K-means are always located at the corresponding mode of their local density, we propose to shift the SVDD ball to the centroid of the data such that it may fit better with the distribution of the data in a cluster. Our new objective function is then formulated as, \footnote{We choose the squared L2 norm distance as a convenient for optimization. There are also other robust distance such as non-squared L2 norm distance \cite{nie2014optimal}.}
\begin{equation}
\begin{split}
 &min_{R,\xi_{i} }  \ R^2+\lambda\sum_{i=1}^{N}\xi_{i} \\
 &s.t. \ \Arrowvert x_{i}-a \Arrowvert^2 \le R^2+\xi_{i} \\
 & ~~~~~~ a=\frac{1}{N}\sum_{i=1}^{N}x_{i}\\
 & ~~~~~~ \xi_{i} \ge 0,
\end{split}\label{eq_csvdd}
\end{equation}
and its Lagrange function is as follows,
\begin{equation}
\begin{split}
&L(R,\xi,\alpha,\beta)=R^2+\lambda\sum_{i=1}^{N}\xi_{i}+\sum_{i=1}^{N}\alpha_{i}\{\Arrowvert x_{i}-a \Arrowvert^2-R^2-\xi_{i}\}\\
&~~~~~~~~~~~~~~~~~~~~~~-\sum_{i=1}^{N}\beta_{i}\xi_{i},
\end{split}\label{lagf}
\end{equation}
where $\alpha_{i}\ge0$ and $\beta_{i}\ge0$ are the corresponding Lagrange multipliers. According to KKT Conditions, we have,

\begin{equation}
\begin{split}
\frac{\partial L}{\partial R}=2R-2R\sum_{i=1}^{N}\alpha_{i}=0~,~\sum_{i=1}^{N}\alpha_{i}=1\\
\end{split}\label{stalpha}
\end{equation}
\begin{equation}
\begin{split}
\frac{\partial L}{\partial \xi_{i}}=\lambda-\alpha_{i}-\beta_{i}=0
\end{split}\label{stxi}
\end{equation}
Taking Eq.(\ref{stalpha}) and Eq.(\ref{stxi}) into the Lagrange function (\ref{lagf}) we get that
\begin{displaymath}
L(R,\xi,\alpha,\beta)=\sum_{i=1}^{N}\alpha_{i}\{\Arrowvert x_{i}-a \Arrowvert^2\}.
\end{displaymath}
Recalling that $a=\frac{1}{N}\sum_{i=1}^{N}x_{i}$, one has the following dual function,
\begin{equation}
\begin{split}
 &max \ \sum_{i}\alpha_{i} \langle x_{i},x_{i} \rangle-\frac{2}{N}\sum_{i}\sum_{j}\alpha_{i}\langle x_{i},x_{j}\rangle  \\
 &s.t. \ \sum_{i}\alpha_{i}=1 \ , \ \alpha_{i}\in[0,\lambda]\ ,\ i=1,...,N.
\end{split} \label{eq_csvdd_dual}
\end{equation}
This can be reformulated as
\begin{equation}
\begin{split}
 &min \  \frac{2}{N}\alpha^THe-\alpha^TF \\
 &s.t. \ \alpha^Te=1 \ , \ \alpha_{i}\in[0,\lambda]\ ,\ i=1,...,N,
\end{split}\label{eq_csvdd_dual_m}
\end{equation}
where $H=(\langle x_{i},x_{j}\rangle))_{N\times N}~,~F=(\langle x_{i},x_{i}\rangle)_{N\times 1}~,~e=(1,1,...,1)^T$ . This objective function is linear to $\alpha$, and thus can be solved efficiently with a linear programming algorithm.

Since the model is centered towards the mode of the distribution of the data points in a cluster, we named our method as C-SVDD (centered-SVDD). Fig.~\ref{fig:LPSVDD} shows the difference between SVDD and C-SVDD, where the left is from SVDD and the right from C-SVDD. We can see that our new model aligns better with the density of the data points, as expected. It is also worth mentioning that the normalization parameter $\lambda$ plays an important role in our model - a larger $\lambda$ value would allow more noise to enter the ball, while $\lambda = 0$, the C-SVDD model actually reduces to the naive single-cluster K-means. More discussions on setting this value empirically will be given in Section~\ref{sec_exps}.

After the model is trained, we use the modified ``triangle" encoding (E.q.~\ref{eq_softsvdd}) for feature encoding, with almost the same computational complexity with its K-means counterpart.

\subsection{K-means Encoding vs. C-SVDD Encoding}

\begin{figure*}[!t]
\centering
\includegraphics[width=0.75\linewidth]{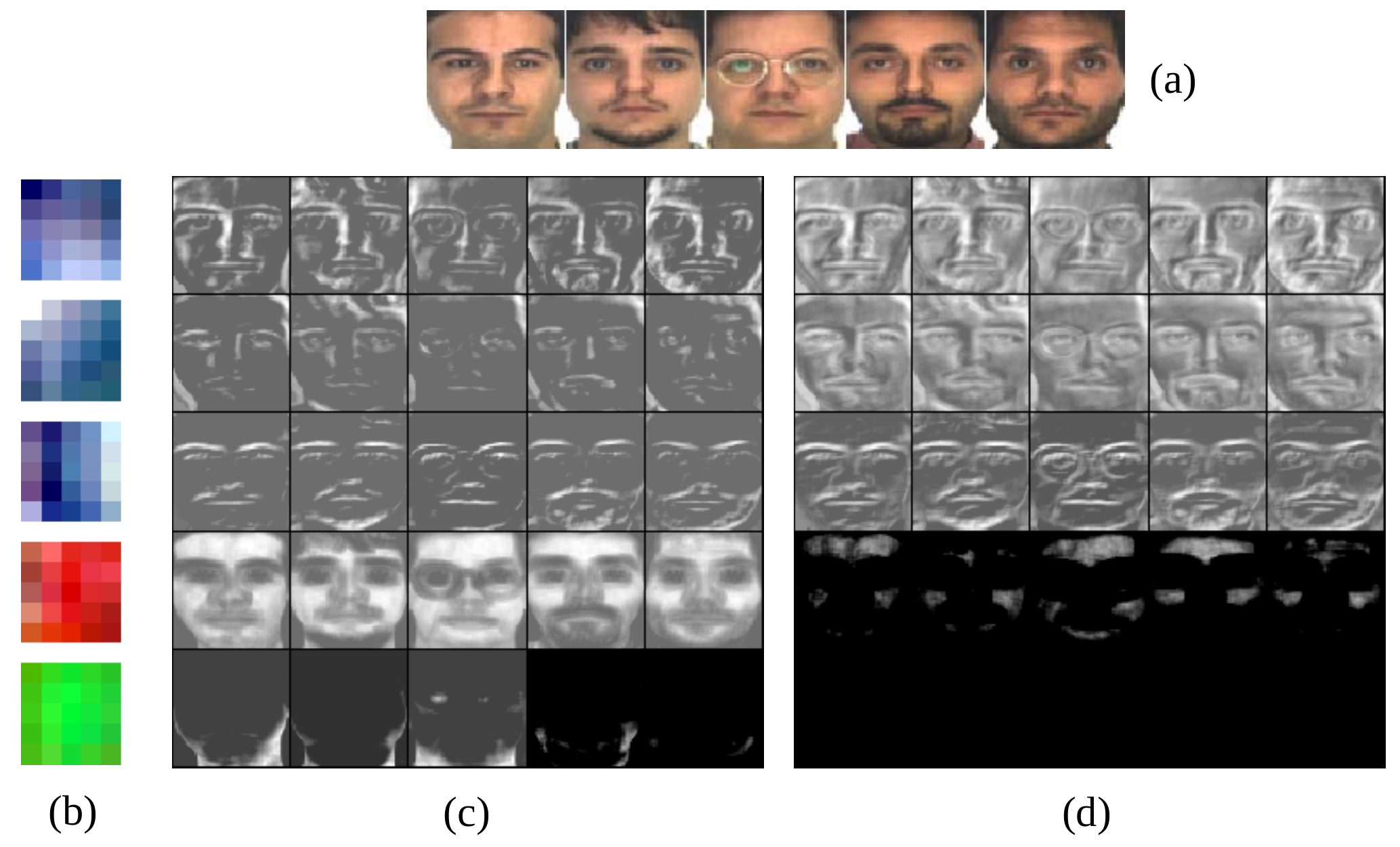}
\caption{Illustration of feature maps of five face images (a) using K-means (c) and C-SVDD (d) respectively, based on five local dictionary atoms (b), where maps in each row are corresponding to one atom next to it while each column corresponding to one face. For the response values in a feature map, the darker the lower.}
\label{fig_faces}
\end{figure*}

\begin{figure*}[!t]
\centering
\subfigure[]{\includegraphics[width=0.312\textwidth]{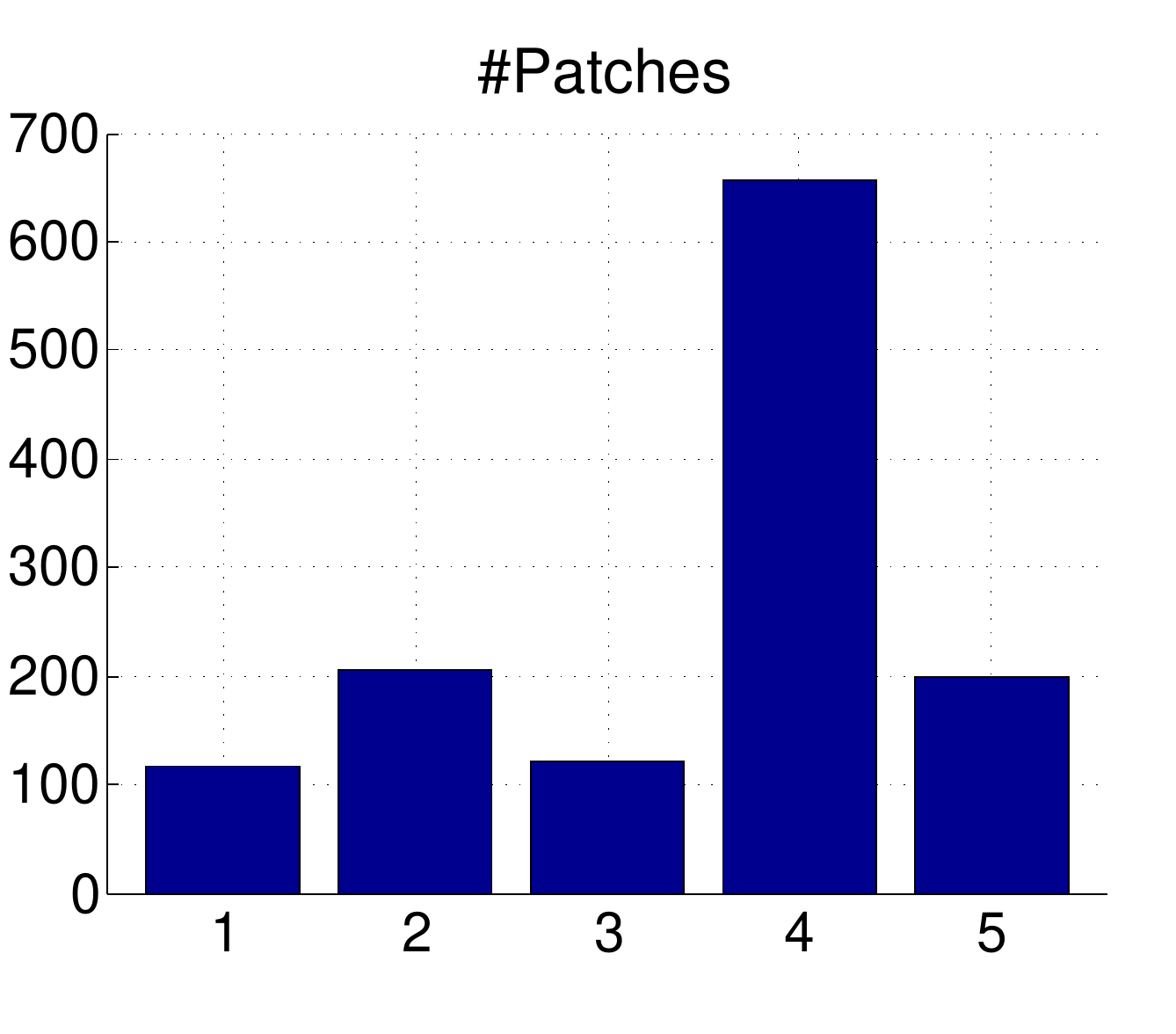}
\label{fig:histp}}
\subfigure[]{\includegraphics[width=0.312\textwidth]{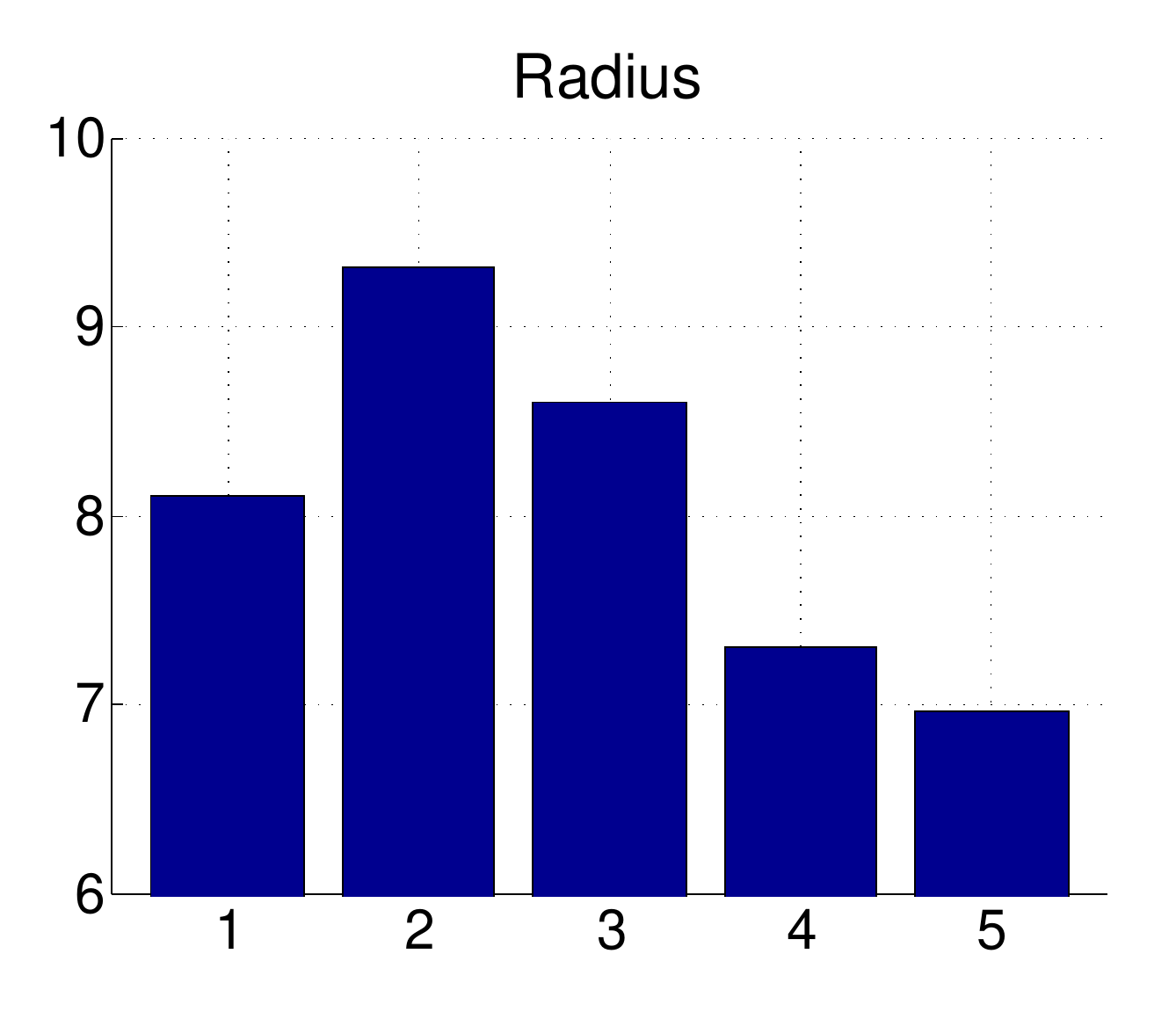}
\label{fig:histr}}
\subfigure[]{\includegraphics[width=0.312\textwidth]{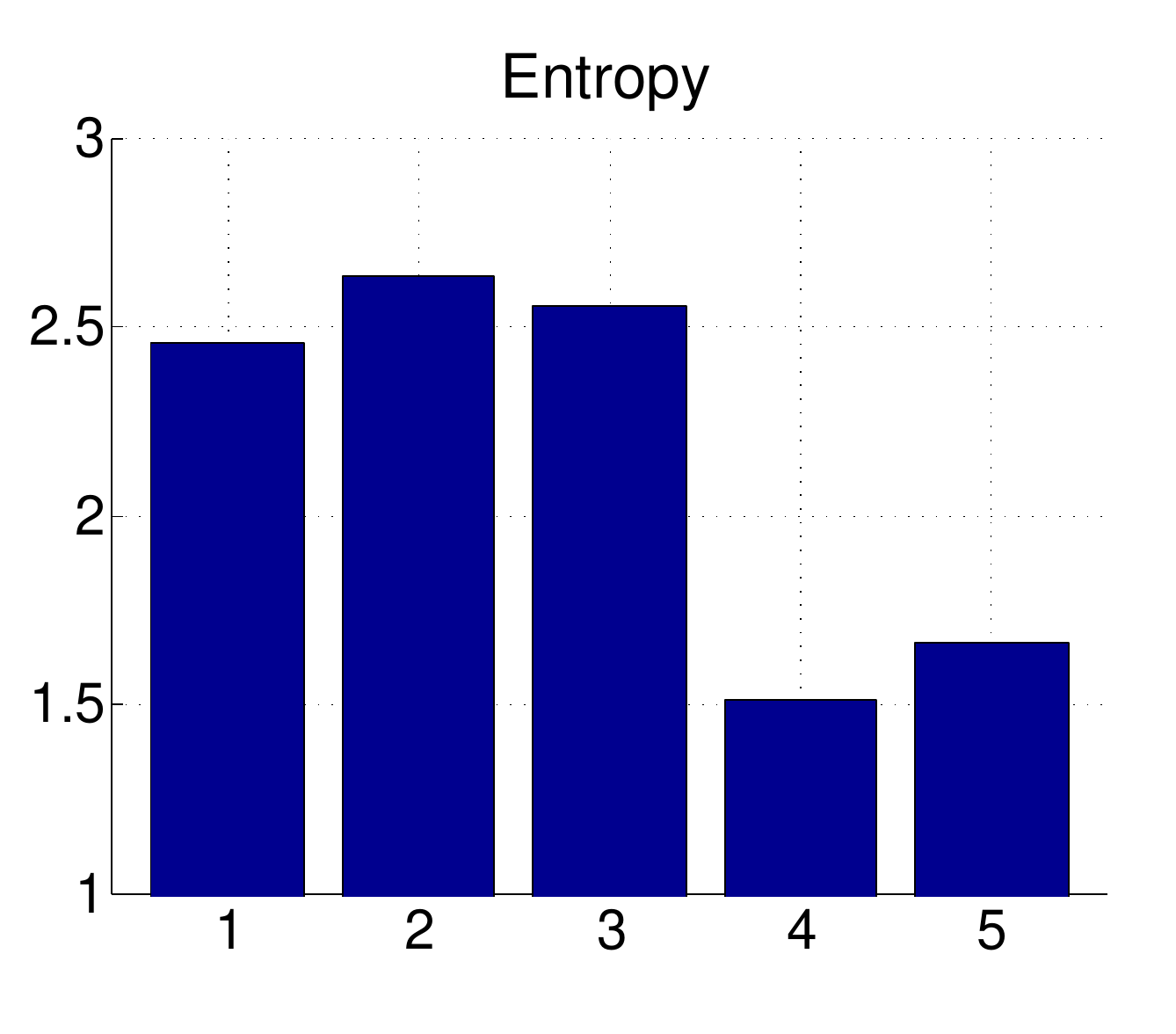}
\label{fig:histe}}
\caption{Distribution of the number of patches attracted by each atom (a), the radius of the corresponding SVDD ball (b), and the entropy (c) over the five atoms shown in Fig.\ref{fig_faces} (leftmost) }
\label{fig_hist}
\end{figure*}

To this end, it will be useful to take a brief discussion on the difference of two kinds of feature maps, i.e., K-means-based ``triangle" encoding (E.q.~\ref{eq_soft}) and our C-SVDD-based one \footnote{Hereinafter we will call them respectively ``K-means encoding" and ``C-SVDD encoding" for short without confusion.}. For this a pilot experiment is conducted. Particularly, we learn a very small dictionary containing only five atoms using five face images, by clustering ZCA-whitened patches randomly sampled from the faces, and then take these for feature encoding. Fig.\ref{fig_faces} illustrates the face images used for dictionary learning (top) and the five learnt atoms (leftmost). The feature maps of face images encoded by the K-means encoding method and those by the C-SVDD encoding method are respectively shown in Fig.\ref{fig_faces} (a) and Fig.\ref{fig_faces} (b), where each row \textcolor[rgb]{0.9,0,0}{is} corresponding to one dictionary atom next to it and each column corresponding to one face.

By comparing the feature maps shown in Fig.\ref{fig_faces} (a) and Fig.\ref{fig_faces} (b), one can see that the C-SVDD-based ones contain more detailed information than the K-means feature maps for the first three atoms, while the responses of the last two atoms are largely suppressed by our method (c.f., last two rows of Fig.\ref{fig_faces} (b)). To further understand this phenomenon, we plot the entropy of each atom (by treating them as a small image patch) in Fig.\ref{fig_hist} (c). The figure shows that the entropy of the last two atoms is much smaller than that of the first three ones, which indicates that the local appearance patterns captured by these last two atoms are much simpler than those by the first three. Hence these two atoms will tend to be widely used by many faces, resulting in reduced discriminative capability in distinguishing different subjects. In this sense, it will be useful to suppress their responses (c.f., the last two rows of Fig.\ref{fig_faces} (b)).

It is also useful to inspect the distribution of local facial patches attracted by these atoms. Fig.\ref{fig_hist} (a) gives the results. It can be seen that this distribution is not uniform and the number of local patches attracted by the fourth atom is significantly larger than those by other atoms. As a result, for K-means encoding method, the feature maps yielded by this atom show much more rich details than others (see the fourth row of Fig.\ref{fig_faces} (a)), potentially indicating that it could play more important roles than others in the subsequent classification task. However, as explained above, since this atom actually contains much less information than the first three atoms (low entropy and being a ``common word"), it is really not good to over-emphasize its importance in feature encoding.

This drawback of K-means feature mapping is largely bypassed by our C-SVDD-based scheme. As shown Fig.\ref{fig_hist} (b), the fourth atom actually represents a very small cluster. In fact, the radius of C-SVDD ball corresponding to the more informative atom tends to be large, and one major advantage of our C-SVDD-based strategy is that it is capable to exploit this characteristic of dictionary atoms for more effective feature encoding, as shown in the first three rows of Fig.\ref{fig_faces} (b). This partially explains the superior performance of the proposed C-SVDD method compared to its K-means counterpart (c.f., experimental results in Section~\ref{sec_exps}).

\begin{figure*}[!t]
\centering
\subfigure[]{\includegraphics[width=0.31\textwidth]{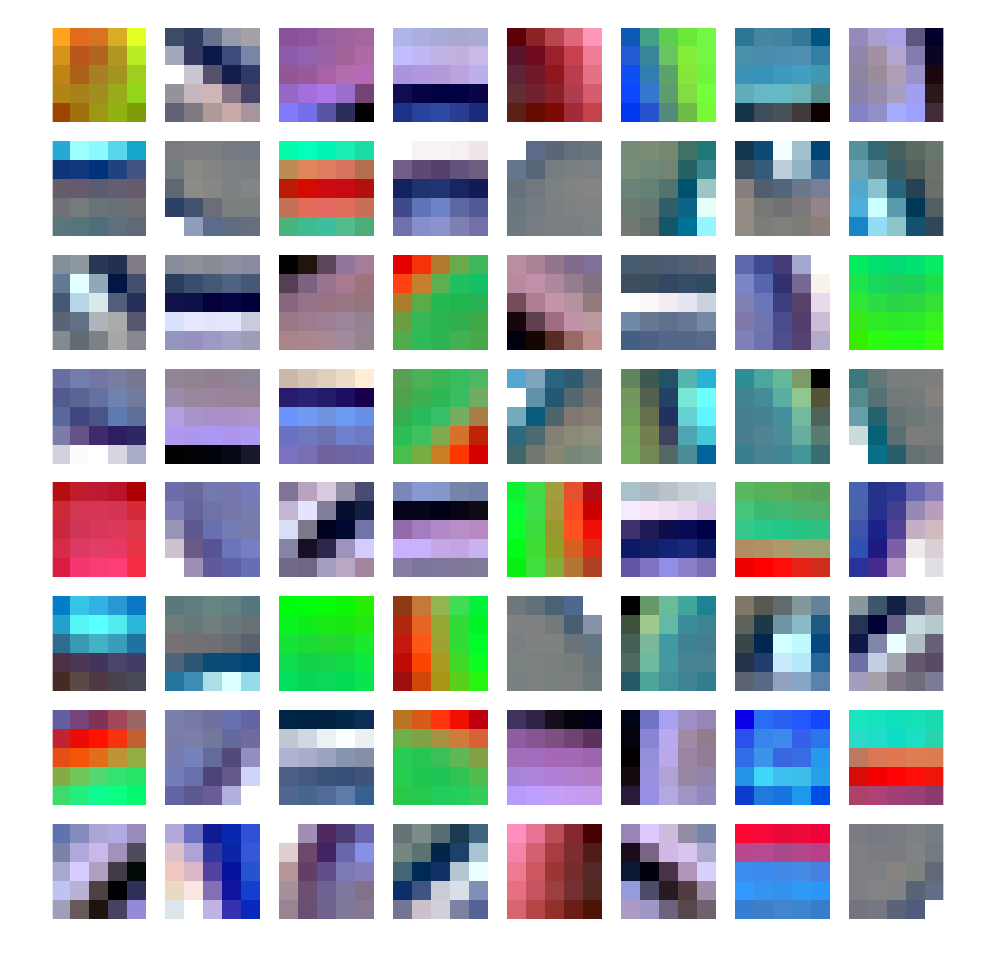}
\label{fig:rf5}}
\subfigure[]{\includegraphics[width=0.31\textwidth]{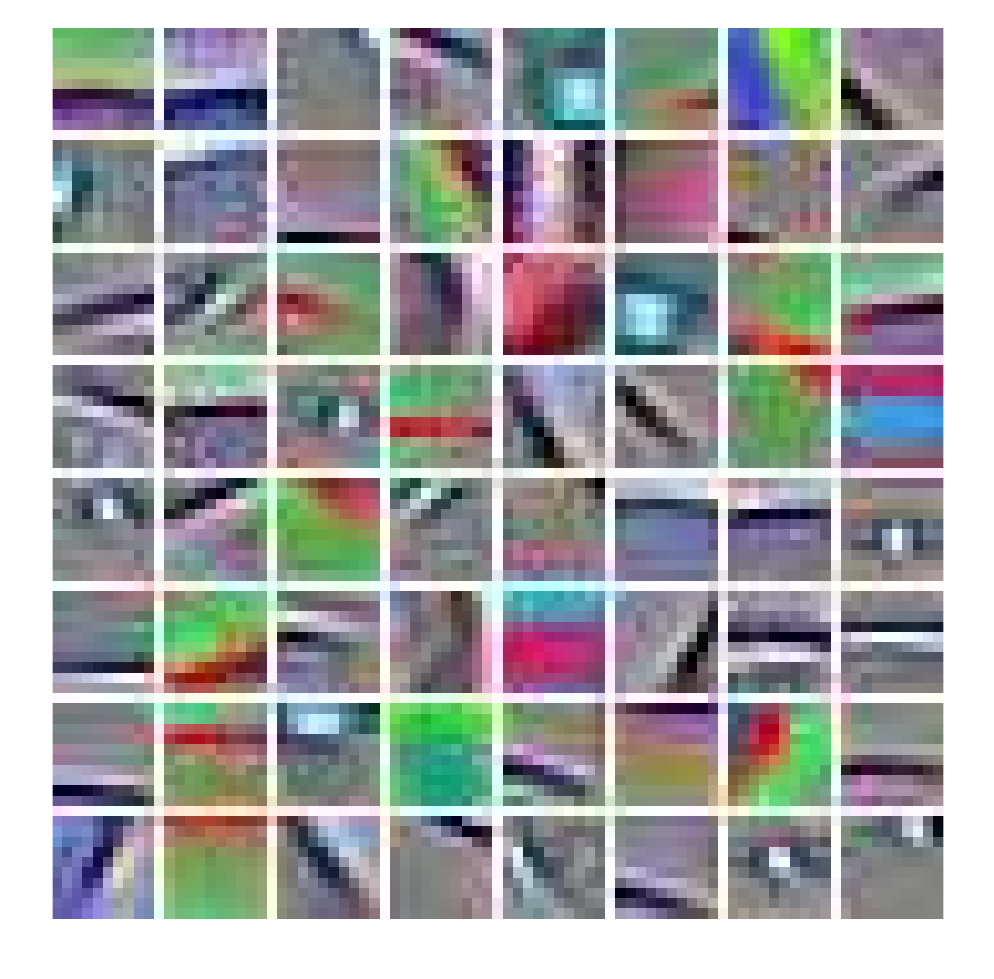}
\label{fig:rf9}}
\subfigure[]{\includegraphics[width=0.325\textwidth]{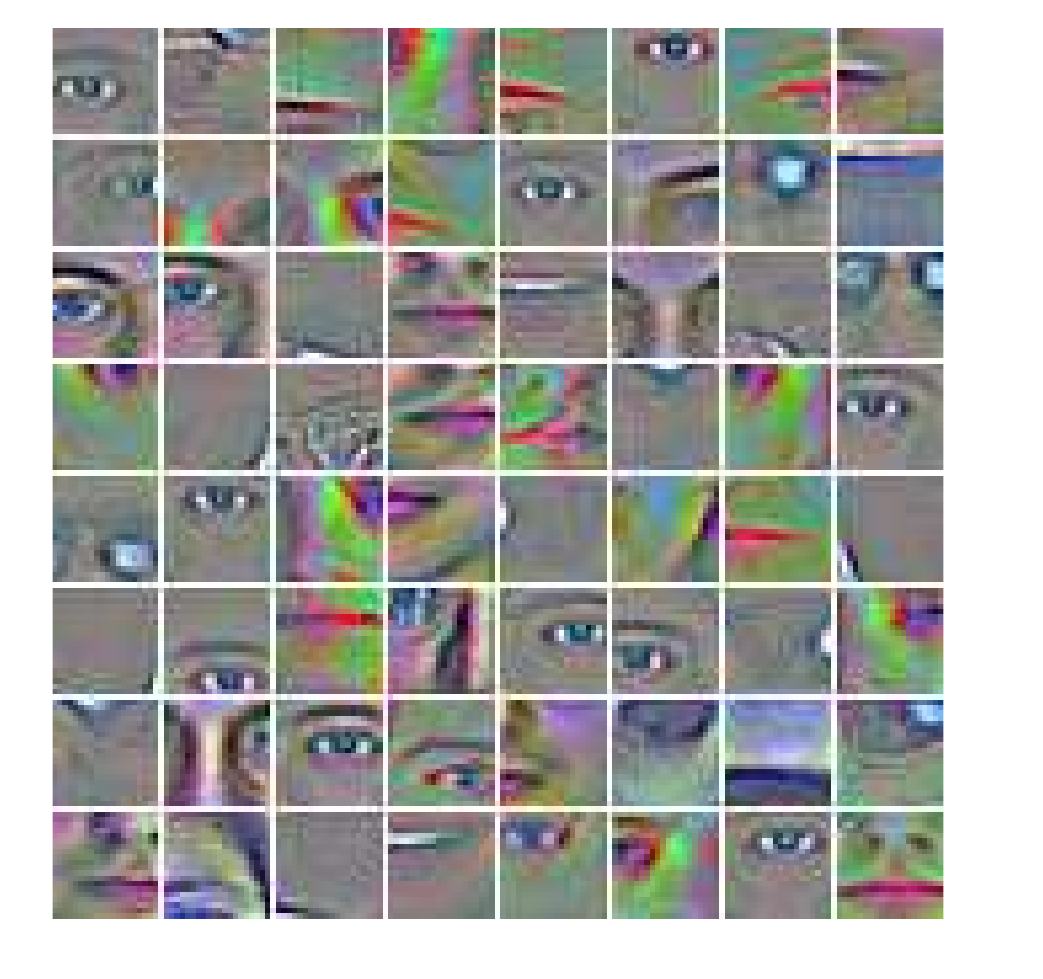}
\label{fig:rf20}}
\caption{Features of different scales learnt from face images. The size of original face images is $64 \times 64$ in pixels. (a) size $=$ 5 (b) size $=$ 10 (c) size $=$ 20}
\label{fig:msfea}
\end{figure*}

\subsection{Encoding Feature Maps with SIFT Representation}
 Traditional unsupervised methods like Bag of Words (BoW) model \cite{csurka2004visual} usually generate a global feature representation by simply histogramming over local codings, while ignoring spatial relationship between local patches.

 One problem of preserving spatial information in feature representation is due to the huge dimension of feature maps. Suppose that the size of a receptive field is $r \times r$, and the size of an input image is $D \times D$. After densely extracting patches and encoding them, we would obtain $K$ feature maps, one for each filter, with each of size $S \times S$ ($S = D-r+1$). Particularly, for small images with $D = 96$, and a small dictionary with size $K = 256$ and with size of its filter $r = 5$, the resulting dimension of $K$ feature maps is nearly $2M$, which is too large for many applications. One can use such methods as average pooling or max pooling to reduce the size of feature maps. For  $p \times p$ sized pooling blocks, the size of a feature map can be reduced to $\lceil \frac{S}{p} \rceil \times \lceil \frac{S}{p} \rceil$. In the above example if we set $p = 5$, the dimension of each map becomes $19 \times 19 = 361$, which is still too big when concatenating $K$ maps. However, if we choose a bigger pooling window, the more spatial information will be lost.

In this paper we proposed a variant of SIFT-representation to address the above issues. SIFT is a widely used descriptor in computer vision and is helpful to suppress the noise and improve the invariant properties of the final feature representation. The way we get SIFT representation is not in general way which extracts a 128-bit SIFT-descriptors densely. This will also cause to a very high dimensionality. For example, if we extract 128 dimensional SIFT-descriptors densely in $256$ feature maps with the size of $16\times 16$ in pixel, the dimension of the obtained representation vector will be as high as over $11.8$M ($250\times 19\times 19\times 128 =11,829,248$). To address this issue, we first divide each feature map into $m \times m$ blocks and then only extract an 8-bit gradient histogram from each block in the same way as SIFT does. This results in a feature representation with dimension of $m\times m\times 8$ for each map (Such as if m=3, then the dim is only 72bit). In this way we significantly reduce the dimensionality while preserving rich information for the subsequent task.

\subsection{Multi-scale Receptive Field Voting}
Next we extend our method to exploit multi-scale information for better feature learning. A multi-scale method is a way to describe the objects of interest in different sizes of context. This would be useful since patches of a fixed size can seldom characterize an object well - actually they can only capture local appearance information limited in that size. For example, if the size is very small, information about edges could be captured but the information on how to combine these into more meaningful patterns such as motifs, parts, poselets, and object, is lost, while information about these entities at different levels is valuable in that they are not only discriminative by itself but complementary to each other as well. Most popular manually designed feature descriptors, such as SIFT or HoG, address this problem to some extend by pooling image gradients into edglets-like features, but it is still unclear, for example, how to assemble edglets into motifs using these methods. Convolutional neural network provides a simple and comprehensive solution to this issue by automatically learn hierarchies of features ranging from edglets to objects. However, during this procedure, information on where those high-level patterns are found becomes more and more ambiguous.

Because our C-SVDDNet is a single-layer network,  it is difficult to learn multi-scale information in a hierarchical way. Instead, we take a naive way to obtain multi-scale information by using receptive fields of different sizes. In particular, we fetch patches with $S_i \times S_i, i=1,2,3$ squares in size from training images and use these to train dictionary atoms with corresponding size through K-means. Fig.\ref{fig:msfea} shows some examples of atoms we learnt on a face dataset. One can see that these feature extractors are similar to those learnt using a typical ConvNet. Specifically, with the increasing window size, the learnt features become more understandable - for example, as shown in Fig.\ref{fig:msfea} (c), using a receptive field with size of $20 \times 20$ on face images of $64 \times 64$, we successfully learned facial parts such as the eyes, the mouth, and so on, while a smaller receptive field gives us some oriented filters, as shown in Fig.\ref{fig:msfea} (a). At each scale we train several networks with different pooling window. One advantage of this method is that it is very efficient to learn and is effective in capture salient features in a multi-scale context. However, it will not tell us how the bigger patterns are explained by smaller ones - such information would be useful from a generative angle.

To use the learnt multi-scale information for classification, we train a separate classifier on the output layer of the corresponding network (view) according to different receptive sizes and different pooling sizes, then combine them under a boosting framework. Particularly, assume that the total number of categories is $C$, and we have $M$ scales (with $K$ different number of pooling sizes for each scale), then we have to learn $M \times K\times C$ output nodes. These nodes are corresponding to $M\times K$ multi-class classifiers. Let us denote the parameter of the $t-$ th classifier $\theta_{t}\in R^{D\times C}$ ($D$ is the dimension of feature representation) as $\theta_{t} = [w_{t1},w_{t2},...,w_{tC}]$, where $w_{tk}$ is the weight vector for the $k$-th category. We first train these parameters using a series of one-versus-rest $L_2$-SVM classifiers, and then normalize the outputs of each classifier using a soft max function,
\begin{equation}\label{eq_softmax}
  f_{tk}(x_{i})= \frac{exp(w_{tk}^Tx_{i})}{\sum^C_{c=1}exp(w_{tc}^Tx_{i}).}
\end{equation}

Finally, the normalized predictions $f_{tk}$ are combined to make the final decision,
\begin{equation}\label{eq_softmax}
  g(x_{i})= argmax_c \sum_{t} a_{tc}^T f_{t}(x_i).
\end{equation}
where $f_t=\{f_{t1},f_{t2},...,f_{tC}\}$ is the output vector of the $t$-th classifier, and the corresponding combination coefficients $a_{tc}$ are trained using the following objective,
\begin{equation}\label{eq_msmv}
 min_{a_c} ~ \sum_{i}max(0,1- \sum_{t} a_{tc}^T f_{t}(x_i))^2 + \lambda ||a_c||_2
\end{equation}
This is the same type of one-versus-rest $L_2$-SVM mentioned before.


\section{Experiments AND Analysis}\label{sec_exps}

To evaluate the performance of the proposed C-SVDDNet, we conduct extensive experiments on four datasets including two object classification datasets(STL-10 \cite{coates2010analysis}, MINST \cite{lecun1998gradient}) and two image retrieval datasets(Holiday \cite{jegou2008hamming}, INRIA Copydays \cite{douze2009evaluation}).

\subsection{Experiment Settings}

All the images undergo whitening preprocessing before feeding them into the network. The whitening operation linearly transforms the data such that their covariance matrix becomes unit sphere, hence justifying the Euclidean distance we use in the K-means clustering procedure.

Unless otherwise noted, the parameter settings listed in Table.\ref{tb:para} apply to all experiments. The influence of some important parameters, such as the number of filters, will be investigated in more detail in the subsequent sections. For single scale network the receptive field is set to be $5 \times 5$ by default across all the datasets, as recommended in \cite{coates2010analysis}, while in multi-scale version, we use receptive fields in three scales, as shown in Table.\ref{tb:para}.

For C-SVDD ball there is a regularization parameter $\lambda$ to set. This parameter allows us to control the amount of noise we are willing to tolerant to.
As can be seen from E.q.\ref{fig:unequal}, a small $\lambda$ value encourages a tight ball. We set $\lambda=1$ by default for most datasets except for those with too noisy background are set to 0.005. Furthermore, the centers in C-SVDD are set as the same as those in k-means, so that we can safely ignore the effect of the initialization of k-means.

Throughout the experiments, we use Coates' K-means ``triangle" encoding method \cite{coates2010analysis} (c.f., Section~\ref{subsec_softcoding}) as baseline (denoted as `K-means'), while its direct counterpart method by simply replacing ``triangle" encoding with C-SVDD encoding is denoted as `C-SVDD'. Furthermore, we denote the proposed single layer network as `C-SVDDNet', and its multi-scale version as `MSRV + C-SVDDNet'. In addition, we re-evaluate the baseline method \cite{coates2010analysis} within the proposed network by replacing its component of C-SVDD with the K-means-based encoding, denoted as `K-meansNet'.

\begin{table}[t!]
\caption{DEFAULT PARAMETER SETTINGS FOR OUR METHODS.}
\begin{center}
\begin{tabular}{c|c}
\hline
Parameter & value  \\
\hline
\hline
\#clusters  &$\le$500  \\
\hline
Size of receptive field &$5\times 5^*, 7\times 7, 9\times 9$\\
\hline
size of average pooling &$4\times 4^*, 1\times 1, 3\times 3$\\
\hline
$\lambda$ of C-SVDD  &$1^*, 0.005$\\
\hline
\multicolumn{2}{l}{*-default setting for the non-multi-scale network.}
\end{tabular}
\end{center}
\label{tb:para}
\end{table}


\subsection{Analysis of the Proposed Method}
First of all we conduct extensive experiments on the STL-10 dataset to investigate the behavior of the proposed method. The STL-10 is a large image dataset popularly used to evaluate algorithms of unsupervised feature learning or self-taught learning. Besides 100,000 unlabeled images, it contains 13,000 labeled images from 10 object classes, among which 5,000 images are partitioned for training while the remaining 8,000 images for testing. All the images are color images with $96 \times 96$ pixels in size. There are 10 pre-defined overlapped folds of training images, with 1000 images in each fold. In each fold, a classifier is trained on a set of 1000 training images, and tested on all 8000 testing images. In consistence with  \cite{coates2010analysis}, we report the average accuracy across 10 folds. For unsupervised feature learning we randomly select 20,000 unlabeled data. The size of spatial pooling is $4 \times 4$, hence the size of feature maps fed for SIFT representation is $23 \times 23$. For multi-scale receptive voting we use 2 scale ($5 \times 5$ and $7 \times 7$), on each of which we perform spatial pooling in 5 sizes ranging from $2 \times 2$ to $6 \times 6$.

\def\para#1{\medskip\noindent{\bf #1}}

\para{Do we really need a large number of local features?}
\begin{figure}[t!]
\centering
\includegraphics[width=0.95\linewidth]{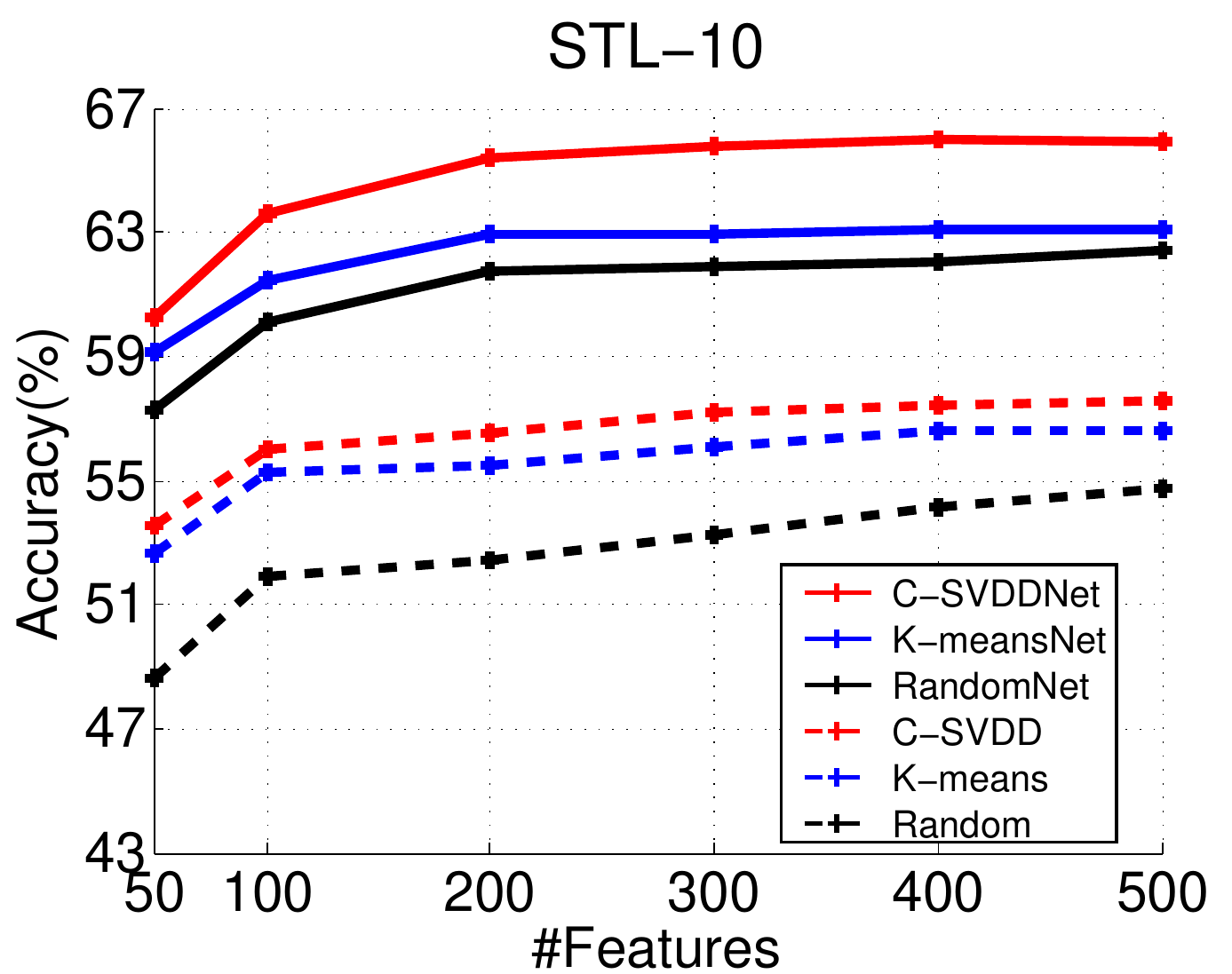}
\caption{The effect of different number of features on the performance with different methods on the STL-10 dataset. All parameters here (such as pooling size) are set as the same as in Fig.\ref{fig:barpl}. }
\label{fig:stlfeature}
\end{figure}
By the number of features, we mean the number of filters $K$ used for feature extraction, which is equal to the number of dictionary atoms. One of the major conclusions of Coates et al.'s series of controlled experiments on single layer unsupervised feature learning network \cite{coates2010analysis} is that compared to the choice of particular learning algorithm, the parameters that define the feature extraction pipeline, especially the number of features, have much more deep impact on the performance. Using a K-means network with 4000 features, for example, they are able to achieve surprisingly good performance on several benchmark datasets - even better than those with much deeper architectures such as Deep Boltzmann Machine \cite{salakhutdinov2009deep} and Sparse Auto-encoder \cite{coates2010analysis}.

However, one drawback accompanying this large dictionary is that a very crude pooling size has to be adopted (e.g., $46\times 46$ over $92\times 92$ feature maps) to condense the resulting feature maps, otherwise the dimensionality of the final feature representation could be prohibitively high. For example, a $3\times 3$ pooling over 4000 feature maps with $92\times 92$ in size would lead to a total number of features over 3.8M. Hence the first question we investigate is that whether such a large number of features are really needed all the time?

Fig.\ref{fig:stlfeature} gives the performance curves according to varying number of features with different methods on the STL-10 dataset. Besides the aforementioned methods, in this figure we also give the results of random dictionary (i.e, local dictionary atoms are obtained randomly without being fine tuned by k-means, denoted as ``Random" ) and of the combination of random dictionary and SIFT representation (denoted as  'RandomNet').

It can be seen that with the increasing number of features, the performance of both K-means and C-SVDD methods rises, which is consistent with the results by Coates et al. \cite{coates2010analysis}. One possible explanation is that since both K-means encoding and C-SVDD encoding use the learnt dictionary to extract non-linear features, more dictionary atoms help to disentangle factors of variations in images. In our opinion the capability to learn a large number of atoms at relatively low computational cost is one of the major advantages of K-means based methods for unsupervised feature learning over other algorithms such as Gaussian Mixture Model (GMM), sparse coding, and RBM. For example, it is difficult for a GMM to learn a dictionary with over 800 atoms \cite{coates2010analysis}.

On the other hand, a too large dictionary can increase the redundancy and decrease the efficiency. Hence it is desirable to reduce the number of features while not hurting the performance too much. Fig.\ref{fig:stlfeature} shows that our C-SVDD encoding method consistently works better than the K-means encoding at different number of features, and combining C-SVDD encoding and SIFT-based representation dramatically reduces the needs for large dictionary without scarifying the performance. Actually, Table.\ref{tb:stl} and fig.\ref{fig:stlfeature} show that using our C-SVDD encoding and the SIFT feature representation, the dictionary size reduces by 10 times (from 4,800 \cite{coates2011selecting} to 500) while the performance improves by 12\% (from 53.80\% \cite{coates2011selecting} to 65.92\%).

As for the random dictionary (denoted as 'Random' and 'RandomNet' in Fig.\ref{fig:stlfeature}), it is interesting to see that when the number of atoms is small, random atoms perform much worse than those finetuned by k-means. But as the size of dictionary increases, the performance difference between the random dictionary and K-means dictionary begins to reduce. For example, at 500 features, using random atoms gives a performance of 54.77\%, slightly worse than that of k-means (56.63\%), and the performance of RandomNet (62.45\%) is also close to that of K-meansNet (63.07\%). However the performance of both random methods is all much lower than that of the C-SVDD based methods.


\para{Effect of the pooling size}
\begin{figure}[t!]
\centering
\includegraphics[width=0.95\linewidth]{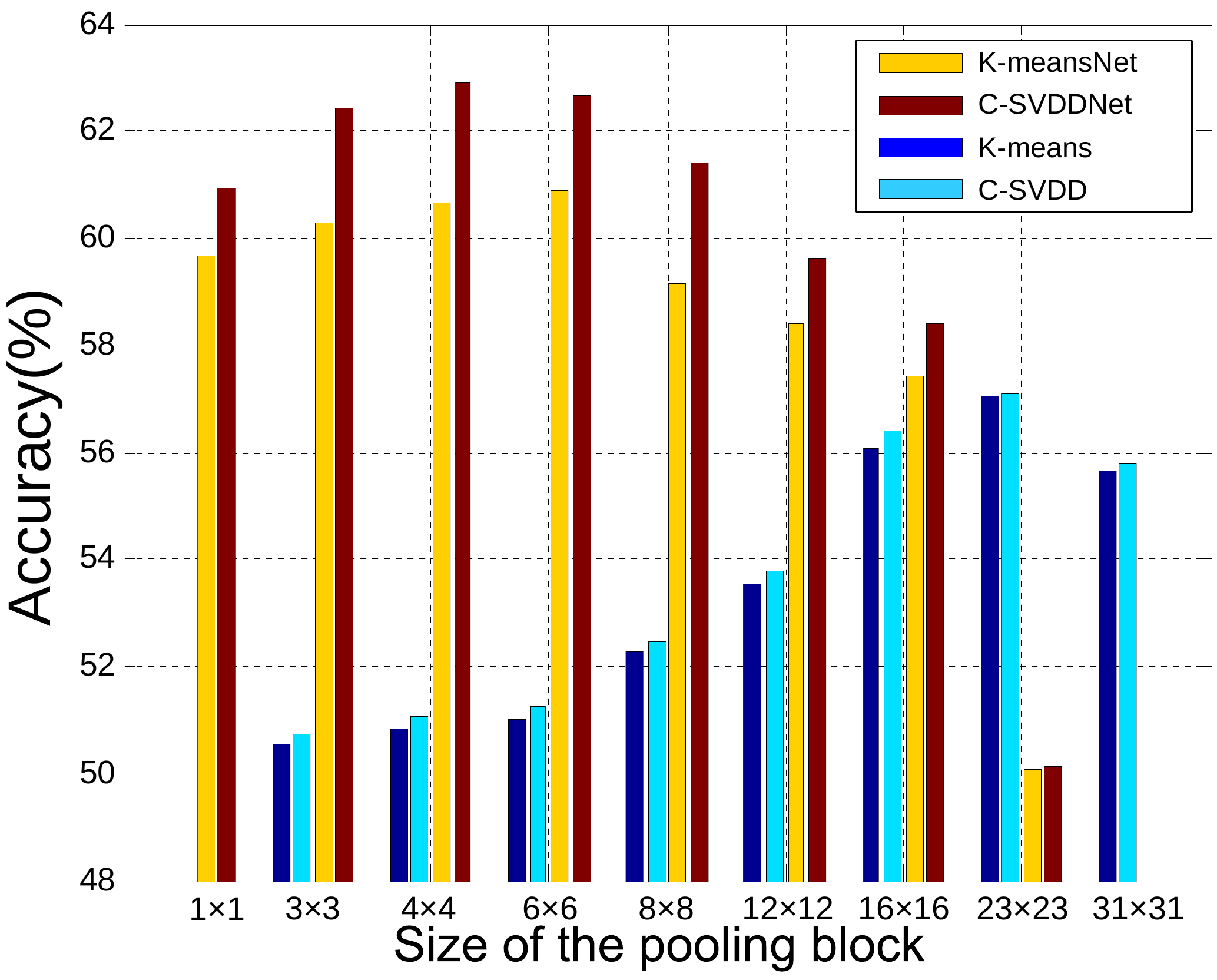}
\caption{The effect of different pooling sizes on the performance with the proposed method on the STL-10 dataset.}
\label{fig:barpl}
\end{figure}
To investigate the effect of different pooling sizes on the performance using the proposed method, we conduct a series of experiments on the STL-10 dataset. Particularly, for a 96 $\times$ 96 original image, we use a receptive field of 5 $\times$ 5 in pixel for feature extraction and obtain a layer of feature maps with 92 $\times$ 92. The pooling blocks are set to be $m\times m$ such that the size of final feature maps after pooling is $\frac{92}{m}$ $\times$ $\frac{92}{m}$. We vary $m\times m$ from 1 $\times$ 1 to 31 $\times$ 31 and record the yielded accuracy. Fig.\ref{fig:barpl} gives the results under different settings. We can see from the figure that generally for the one layer K-means-based network we need bigger block sizes for improved translation invariance, but adding a robust SIFT encoding layer after pooling effectively reduces the needs for large pooling size while obtaining better performance. One possible reason is that this tends to characterize more detailed information of the objects to be represented.

\para{Effect of the multi-scale receptive field voting}
\begin{figure}[t!]
\centering
\includegraphics[width=0.9\linewidth]{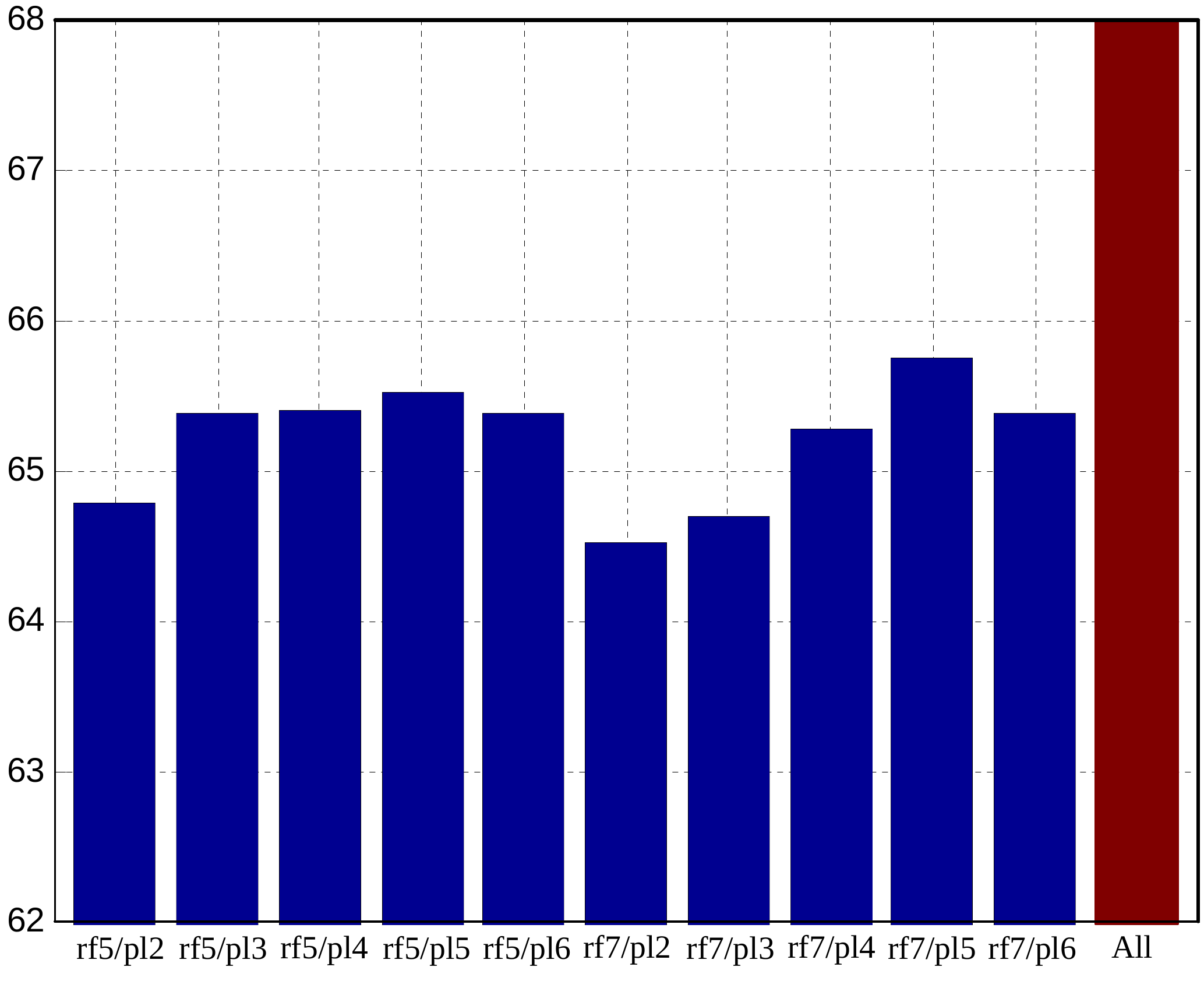}
\caption{Detailed performance of 10 different representations and their ensemble on the STL-10 dataset. These representations are obtained by combining different receptive field size (rf$s$) and pooling size (pl$s$), where rf$s$ indicates a receptive field of $s\times s$, and pl$m$ denotes a pooling block of $m\times m$ in pixel.}
\label{fig:bar}
\end{figure}
Fig.\ref{fig:bar} gives the detailed accuracy of 10 representations using 2 sizes of receptive fields and 5 sizes of pooling blocks. One can see that different representation leads to different prediction accuracy but combining them leads to better performance. This shows that the representations captured with different receptive fields and pooling sizes are complementary to each other.

%

\para{Contribution of components}
\begin{figure}[t!]
\centering
\includegraphics[width=0.8\linewidth]{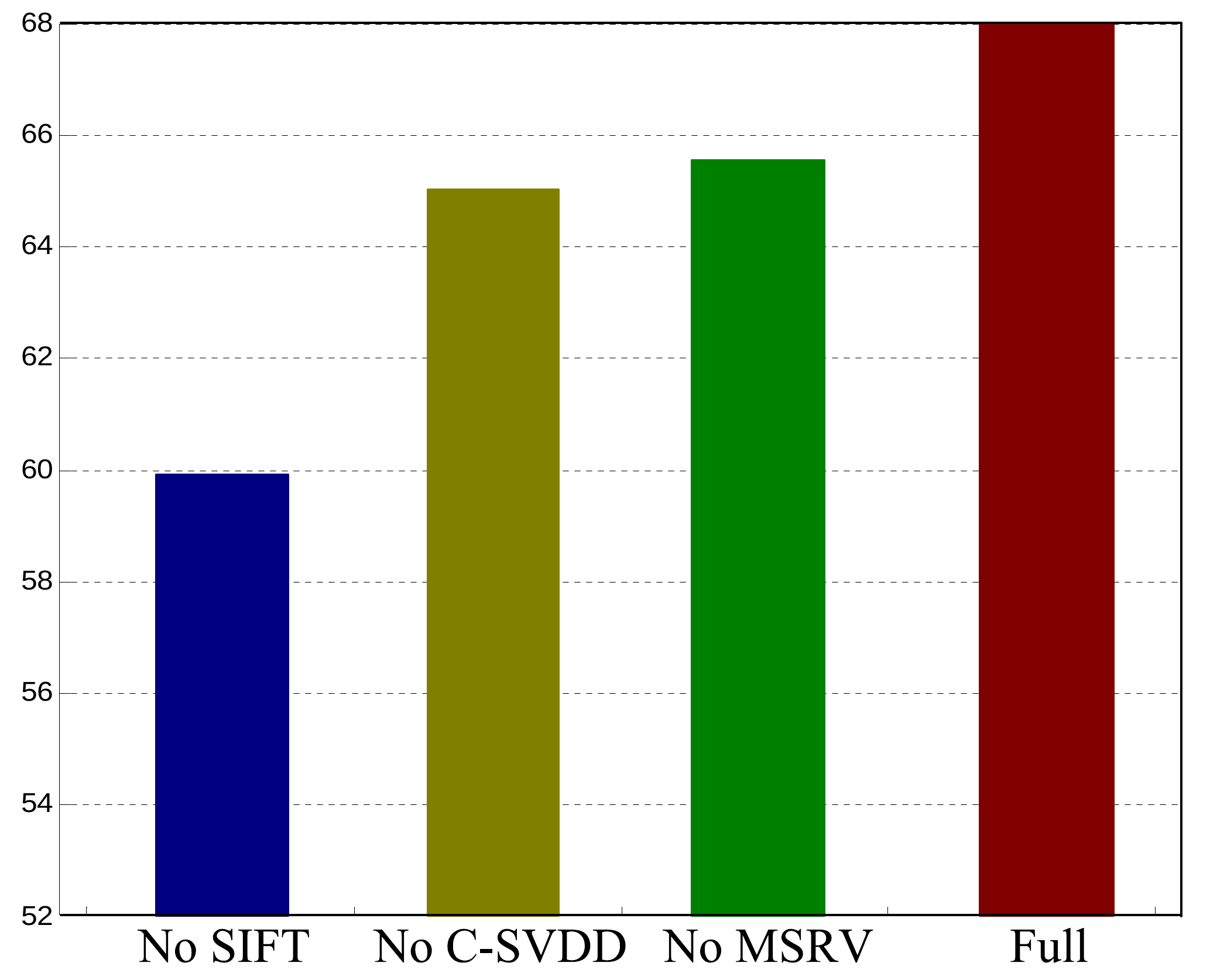}
\caption{The contribution of the three major components of the proposed method to the performance.}
\label{fig:msvbar}
\end{figure}
To illustrate the contributions of the individual stages of the proposed method (i.e., C-SVDD-based encoding, SIFT-representation and multi-scale voting), we conduct a series of experiments on the STL-10 dataset by removing each of the three main stages in turn while leaving the remaining stages in place (the comparison is thus against our full method). Fig.\ref{fig:msvbar} gives the results. In general each stage is beneficial and (not shown) the results are cumulative over the stages, but the SIFT stage seems to contribute most to the performance improvement.  This suggests that taking spatial information into global representation is of importance.

\subsection{Object Classification}
\begin{table}[t!]
\caption{Comparative performance (\%) on the STL-10 dataset.}
\begin{center}
\begin{tabular}{l|l}
\hline
Algorithm & Accuracy($\%$)  \\

\hline
\hline
Selective Receptive Fields (3 Layers) \cite{coates2011selecting} (2011) &60.10 $\pm$ 1.0 \\
\hline
\emph{Trans. Invariant} RBM (TIRBM) \cite{sohn2012learning} (2012) &58.70  \\
\hline
\emph{Simulated visual fixation} ConvNet \cite{zou2012deep} (2012) & 61.00 \\
\hline
Discriminative Sum-Prod. Net (DSPN) \cite{gens2012discriminative} (2012) & 62.30 $\pm$ 1.0 \\
\hline
Hierarchical Matching Pursuit (HMP) \cite{bo2013unsupervised} (2013) &64.50 $\pm$ 1.0 \\
\hline
Deep Feedforward Networks \cite{miclut2014committees} (2014) &68.00 $\pm$ 0.55 \\
\hline
\hline
BoW(K = 4800 D = 4800)   &51.50 $\pm$ 0.6 \\
\hline
VLAD(K = 512 D = 40960)  &57.60 $\pm$ 0.6 \\
\hline
FV (K = 256 D = 40960)  &59.10 $\pm$ 0.8 \\
\hline
\hline
K-means (K = 4800 D = 19200) \cite{coates2011selecting} (2011) &53.80 $\pm$ 1.6 \\
\hline
C-SVDD (K = 4800 D = 19200)  &54.60 $\pm$ 1.5 \\
\hline
K-meansNet (K = 500 D = 36000) & 63.07 $\pm$ 0.6\\
\hline
C-SVDDNet (K = 500 D = 36000) &  65.92 $\pm$ 0.6 \\
\hline
MSRV+K-meansNet & 64.96 $\pm$ 0.4\\
\hline
MSRV+C-SVDDNet & \textbf{68.23} $\pm$ 0.5\\
\hline
\end{tabular}
\end{center}
\label{tb:stl}
\end{table}

\begin{table}[t!]
\caption{Comparative performance (\%) on the MINST dataset.}
\begin{center}
\begin{tabular}{l|l}
\hline
Algorithm & Error($\%$)  \\
\hline
\hline
Deep Boltzmann Machines \cite{salakhutdinov2009deep} (2009) &0.95  \\
\hline
Convolutional Deep Belief Networks \cite{lee2009convolutional} (2009) & 0.82 \\
\hline
Multi-column deep neural networks \cite{ciresan2012multi} (2012) & 0.23 \\
\hline
Network in Network \cite{DBLP:journals/corr/LinCY13} (2013) &0.47\\
\hline
Maxout Networks \cite{goodfellow2013maxout} (2013) &0.45 \\
\hline
Regularization of neural networks \cite{wan2013regularization} (2013) & 0.21 \\
\hline
PCANet \cite{chan2014pcanet} (2014)&0.62\\
\hline
Deeply-Supervised Nets \cite{lee2014deeply} (2014) & 0.39 \\
\hline
\hline
K-means (1600 features) &1.01\\
\hline
C-SVDD  (1600 features) &0.99\\
\hline
K-meansNet (400 features) &0.45\\
\hline
C-SVDDNet (400 features) &0.43\\
\hline
MSRV+K-meansNet &0.36\\
\hline
MSRV+C-SVDDNet &\textbf{0.35}\\
\hline
\end{tabular}
\end{center}
\label{tb:minst}
\end{table}

\para{STL-10 dataset}
Table.\ref{tb:stl} gives our results on the STL-10 dataset. The major challenges of this dataset lie in that its images are captured in the wild with cluttered background, objects in various scales and poses. As before, we compared our method with several feature learning methods with state of the art performance. One can see that our one scale C-SVDD network obtains 65.92\% accuracy, using a filtering dictionary of 500 atoms, outperforms several other feature encoding methods, such as Bag of Words (BoW), Vector of Linearly Agregated Descriptors (VLAD), Fisher vector (FV) and other unsupervised deep learning methods (e.g, \emph{Trans. Invariant} RBM (TIRBM) \cite{sohn2012learning},  Selective Receptive Fields (SRF) \cite{coates2011selecting}, and Discriminative Sum-Product Networks (DSPN) \cite{gens2012discriminative}). This also indicates that spatial information preserving using SIFT is indeed useful in unsupervised feature learning. Also note that replacing the proposed C-SVDD encoding with K-means encoding leads to nearly 3.0\% performance loss, while fusing the multi-scale information gives us about 2.3\% improvement in accuracy, exceeding the current best performer \cite{miclut2014committees} on this challenging dataset.

\para{MINST dataset}
The MNIST is one of the most popular datasets in pattern recognition. It consists of grey valued images of handwritten digits between 0 and 9. It has a training set of 60,000 examples, and a test set of 10,000 examples, all of which have been size-normalized and centered in a fixed-size image with 28 $\times$ 28 in pixel. In training we use a dictionary with 400 atoms for feature mapping, and after pooling/subsampling we break each feature map into 9 blocks to extract SIFT features. For multi-scale receptive voting, we use 3 types of receptive fields: $5 \times 5$, $7 \times 7$ and $9 \times 9$. Combined these with two settings for pooling sizes (i.e., $1\times 1$ and $2\times 2$, respectively), 6 different views/representations can be obtained for each image in this dataset.

\begin{figure}[t!]
\centering
\includegraphics[width=0.8\linewidth]{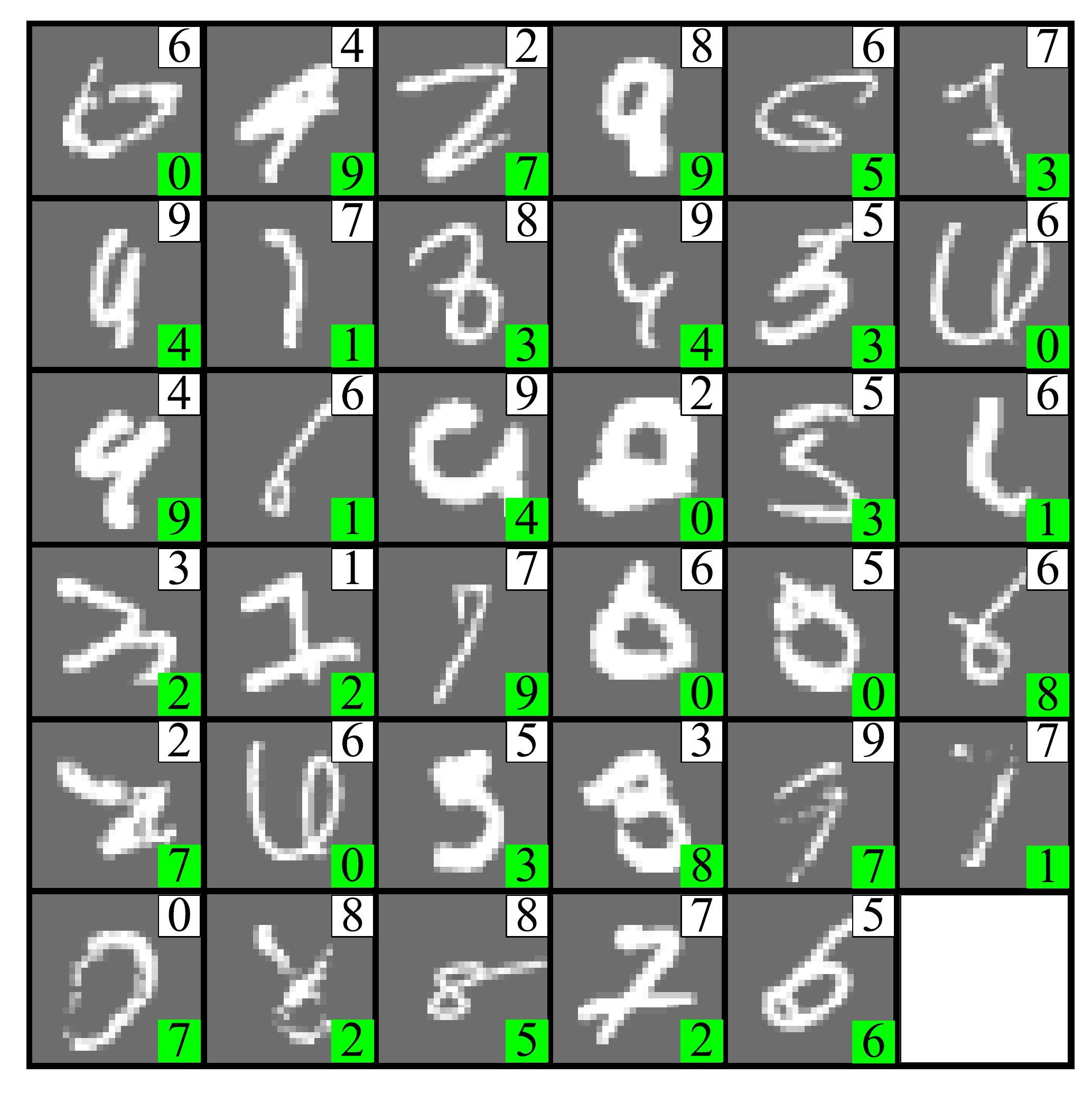}
\caption{All misclassified 35 handwritten digits among 10,000 test examples by our method. The small digit in each white square is the ground truth label of the corresponding image, and the one in the green square is the prediction made by our method.}
\label{fig:minstshow}
\end{figure}

Table.\ref{tb:minst} gives our experimental results on the MINST dataset. It is well-known that deep learning has achieved great success on this task of digit recognition. For example, only 95 among 10,000 test digits are misclassified by the Deep Boltzmann Machines \cite{salakhutdinov2009deep}, while Convolutional Deep Belief Networks \cite{lee2009convolutional} and Maxout Networks \cite{goodfellow2013maxout} respectively reduce this number to 82 and 45. Our simple single layer network (MSRV+C-SVDDnet) achieves an error as low as 0.35$\%$, which is highly competitive to other complex methods using deep architecture. Fig.\ref{fig:minstshow} shows all the 35 misclassified digits by our method, and one can see that these misclassified digits are very confusing even for human beings. Compared to the original K-means network \cite{coates2010analysis}, the proposed method reduces the error rate by 65\%, with much smaller number of filters. This reveals that at least on this dataset with clean background, it is very beneficial to focus more on the representation of the details of the image, rather than emphasizing too much on its global aspects using a large number of filters and a large pooling size.

\subsection{Image Retrieval}

\begin{table*}[t!]
\caption{Comparative performance (mAP \%) on the Holiday dataset.}
\begin{center}
\begin{tabular}{lll|l|l|l|l|l|l}
\hline
Algorithm &K &D  & \multicolumn{6}{|c}{Holidays (mAP \%)}  \\
\hline
   & & & best &D' = 2048 & D' = 512 & D' = 128 & D' = 64 & D' = 32 \\
\hline
\hline
BoW \cite{jegou2012aggregating} (2012)  &20000   &20000  &45.2 &41.80 &44.9 &45.2 &44.4 &41.8\\
\hline
FV \cite{jegou2012aggregating} (2012)  &256     &16384  &62.6 &62.6 &57.0 &53.8 &50.6 &48.6\\
\hline
VLAD \cite{jegou2012aggregating} (2012) &256     &16384  &62.1 &62.1 &56.7 &54.2 &51.3 &48.1\\
\hline
VLAD+adapt+innorm \cite{arandjelovic2013all} (2013) &256 &16384  &64.6 &--- &--- &62.5 &--- &--- \\
\hline
\hline
K-means &3200 &12800   &55.2 &54.5 &54.9 &51.6 &48.3 &44.5\\
\hline
C-SVDD  &3200 &12800  &57.4  &56.8 &57.0 &53.1 &50.5 &46.8\\
\hline
K-meansNet &256 &8192 &62.5 &59.8 &62.5 &61.3 &55.5 &49.5\\
\hline
C-SVDDNet &256 &8192  &\textbf{66.0} &\textbf{63.7} &\textbf{65.8} &\textbf{64.7} &\textbf{59.3} &\textbf{52.1}\\
\hline
MSRV+K-meansNet &256 &8192$\times$8 &66.5 &65.0 &66.3 &65.3 &58.3 &51.5\\
\hline
MSRV+C-SVDDNet &256 &8192$\times$8  &\textbf{70.2} &\textbf{68.6} &\textbf{69.8} &\textbf{68.5} &\textbf{62.5} &\textbf{53.8}\\
\hline
\end{tabular}
\end{center}
\label{tb:holi}
\end{table*}

\begin{table*}[t!]
\caption{Comparative performance (mAP \%) on the Copydays dataset.}
\begin{center}
\begin{tabular}{lll|l|l|l|l}
\hline
Algorithm &K &D  & \multicolumn{2}{|c}{crop 50\%} & \multicolumn{2}{|c}{transformations} \\
\hline
   & & & best &D' = 128 & best &D' = 128\\
\hline
\hline
BoW \cite{jegou2012aggregating} (2012)  &20k   &20k  &100.0 &\textbf{100.0} &54.3 &29.6 \\
\hline
FV \cite{jegou2012aggregating} (2012)   &64    &4096  &98.7 &92.7 &59.6 &41.2 \\
\hline
VLAD \cite{jegou2012aggregating} (2012) &64    &4096  &97.7 &94.2 &59.2 &42.7 \\
\hline
\hline
K-means &3200 &12800   &95.2 &91.5 &47.6 &32.80 \\
\hline
C-SVDD  &3200 &12800  &97.4  &93.8 &52.2 &36.60 \\
\hline
K-meansNet &256 &8192 &96.8 &94.3  &55.4 &37.80 \\
\hline
C-SVDDNet &256 &8192  &\textbf{100.0} &98.1 &\textbf{62.2} &\textbf{52.0} \\
\hline
MSRV+K-meansNet &256 &8192$\times$6 &99.7 &97.9 &58.2 &41.8 \\
\hline
MSRV+C-SVDDNet &256 &8192$\times$6  &\textbf{100.0} &\textbf{100.0} &\textbf{65.6} &\textbf{55.3} \\
\hline
\end{tabular}
\end{center}
\label{tb:copy}
\end{table*}

\para{Holiday dataset} INRIA Holiday dataset consists of 1491 images from personal holiday photos. There are 500 queries, most of which have 1-2 ground truth images. mAP (mean average precision) is employed to measure the retrieval accuracy. We resize all the images to 96 $\times$ 96. In training we use a dictionary with 256 atoms for feature mapping, and after pooling/subsampling we break each feature map into 4 blocks to extract SIFT features. Thus the dimension of final representation is 8196. And we also run PCA for dimensionality reduction as \cite{jegou2012aggregating}. For multi-scale receptive voting, we use 2 types of receptive fields: $5 \times 5$ and $7 \times 7$. Combined these with four settings for pooling sizes (i.e., $3 \times 3$, $4 \times 4$, $5 \times 5$ and $6 \times 6$, respectively), 8 different views/representations can be obtained for each image in this dataset. Note that in image retrieval task, we can not train classifiers so that we just concatenate all the views' representations to combine multi-scale information. In retrieval stage we use Euclidean distance in nearest neighbor searching as in \cite{jegou2012aggregating} and \cite{arandjelovic2013all}, facilitating a fair comparison between various feature representation methods on this task.

Table.\ref{tb:holi} gives our experimental results on this dataset. We compare our method with BOW, VLAD, FV under different dimension ( reduced through PCA). BoW takes a 20k sized filter bank but has the lowest mAP (45.2\%).  Replacing BoW with K-means triangle encoding improves mAP by 10\% (55.2\%), but still needs a large filter bank of 3.2K.

Previous state-of-art unsupervised feature learning methods, i.e.,  VLAD and FV \cite{jegou2012aggregating}, can achieve a high mAP of 62.1\% 62.6\% respectively. And both of them only take a small set of filters of size 256. In \cite{arandjelovic2013all} Arandjelovic combines VLAD with adaptive filter bank and a new normalization to achieve an accuracy of 64.6\%. Our proposed C-SVDDNet can get a mAP of 66.0\% with 256 filters as well. It outperforms VLAD by 3.9\% and VLAD+adapt+innorm by 1.4\%. Even if we reduce its dimension to smaller sizes  with PCA, it consistently achieves the best performance among the compared ones.

Also note that replacing K-means encoding with C-SVDD encoding results in significant improvement (from K-means 55.2\% to C-SVDD 57.4\%, and from K-meansNet 62.5\% to C-SVDDNet 66.0\%). When concatenating multi-scale representation from 8 views, we are able to achieve the highest mAP of 70.2\%, without using any supervision information.


\para{Copydays dataset} INRIA Copydays dataset was designed to evaluate near-duplicate detection [19]. The dataset contains 157 original images. To obtain query images relevant in a copy detection scenario, each image of the dataset has been transformed with three different types of transformation: image resizing, cropping (Here we use only the queries with the cropping parameter fixed to 50\%), strong transformations (print and scan, occlusion, change in contrast, perspective effect, blur, etc). There is in total 229 transformed images, each of which has only a single matching image in the database. All images are resized to 75 $\times$ 75. We use 2 types of receptive fields $5 \times 5$ and $7 \times 7$, together with three pooling sizes (i.e., $3 \times 3$, $4 \times 4$ and $5 \times 5$ respectively), which result 6 different views. To challenge ourself in this experiments we also merge the database with 10k web images as \cite{jegou2012aggregating} does.

It is a large scale retrieval task. Table.\ref{tb:copy} gives our experimental results on this dataset. We can see that in the 50\% cropped circumstance, our C-SVDDNet with only 256 filters can be robust enough to achieve a mAP of 100\% as BoW with 20k filters. When reducing its dimension to 128 bits, it still performs the second best. In the strong transformation setting, our C-SVDDNet achieves a mAP of 62.2\% which outperforms VLAD (59.2\%) and FV (59.6\%) by nearly 3\%.

Furthermore, one can see that C-SVDD encoding allows our C-SVDDNet improve upon K-meansNet by 6.8\% in terms of mAP.  When reduced to 128 bits, our C-SVDDNet achieves a mAP of 52\% under the difficult cases of strong transformation,  which outperforms other compared methods by more than 10\%, while our multi-scale version improves the mAP by 3\%.



\section{Conclusion}\label{sec_conclude}
In this paper, we propose a simple one-layer neural network termed C-SVDDNet for unsupervised feature learning. One of the major advantages of the proposed method is that it allows effective feature representation for many applications, such as object classification and image retrieval, by exploiting unlabeled data which are often cheap and readily available. We show that when properly combined with the SIFT descriptors, such representation could be made even more efficient and discriminant. Extensive experiments on several challenging object classification datasets and image retrieval datasts demonstrate that the proposed method significantly outperforms previous state of the art unsupervised feature learning methods such as Bag of Word, VLAD \cite{jegou2012aggregating}, and FV \cite{sanchez2013image}.

Additionally, we show that for feature representation, a very big dictionary is not necessary, as one could accumulate rich information in each feature map and preserve them with compact encoding (e.g, using the proposed method). This significantly reduces the computational cost. Last but not least, we show that one can use multi-scale information to further improve the performance without training many layers of networks - after all training several shallow networks is much easier than training a deep one.



\section*{Acknowledgements}
The work was financed by the National Science Foundation
of China (61073112), the National Science Foundation of Jiangsu Province (BK2012793), and the Doctoral Fund of Ministry of Education of China (20123218110033).\\

\bibliographystyle{IEEEtran}
\bibliography{refbibtex}

\end{document}